# Explainable AI: A Combined XAI Framework for Explaining Brain Tumour Detection Models

Patrick McGonagle, William Farrelly

Faculty of Engineering and Technology,
Atlantic Technology University
Co. Donegal, Ireland
F92 FC93

Kevin Curran

School of Computing, Engineering & Intelligent Systems, Faculty of Computing, Engineering & Built Environment
Ulster University, Northern Ireland, BT487JL
Corresponding author: kj.curran@ulster.ac.uk

**Abstract**

*This study explores the integration of multiple Explainable AI (XAI) techniques to enhance the interpretability of deep learning models for brain tumour detection. A custom Convolutional Neural Network (CNN) was developed and trained on the BraTS 2021 dataset, achieving 91.24% accuracy in distinguishing between tumour and non-tumour regions. This research combines Gradient-weighted Class Activation Mapping (GRAD-CAM), Layer-wise Relevance Propagation (LRP) and SHapley Additive exPlanations (SHAP) to provide comprehensive insights into the model's decision-making process. This multi-technique approach successfully identified both full and partial tumours, offering layered explanations ranging from broad regions of interest to pixel-level details. GRAD-CAM highlighted important spatial regions, LRP provided detailed pixel-level relevance and SHAP quantified feature contributions. The integrated approach effectively explained model predictions, including cases with partial tumour visibility thus showing superior explanatory power compared to individual XAI methods. This research enhances transparency and trust in AI-driven medical imaging analysis by offering a more comprehensive perspective on the model's reasoning. The study demonstrates the potential of integrated XAI techniques in improving the reliability and interpretability of AI systems in healthcare, particularly for critical tasks like brain tumour detection.*

## 1. Introduction

Medical imaging is a crucial component in modern healthcare, providing essential insights into the human body that can help aid in the diagnosis and treatment of various medical conditions. These techniques have evolved significantly over the years, progressing from basic X-rays to highly sophisticated imaging devices. Today's advanced technologies can capture detailed images of nearly every part of the body with remarkable clarity. Techniques such as X-rays, magnetic resonance imaging (MRI), computed tomography (CT) have revolutionised the field by allowing non-invasive and precise ways of visualising internal structure which allow for a timely and accurate diagnosis (Mshnail et al. 2023). These imaging techniques are vital in the detection and diagnosis of brain tumours, where early and accurate identification can significantly impact patient outcomes.

The integration of artificial intelligence into the analysis of medical imaging has great potential, particularly with deep learning models. AI can automate routine tasks, reduce human error, and efficiently manage the vast amounts of data generated by modern imaging techniques. This improves workflow efficiency, enhances diagnostic accuracy, and enables faster more informed decision-making, which is crucial for early disease detection and personalised treatment (Pinto-Coelho 2023). These deep learning models have the capability to challenge and even surpass the diagnostic accuracy of medical experts. Furthermore, AI models can be trained to recognise subtle patterns that might be missed by the human eye, potentially leading to earlier and more accurate diagnoses (Umer et al. 2021).

However, the 'black box' nature of these models results in a considerable lack of transparency in how they come up with an answer which can lead to mistrust among the users and patients (Chauhan and Sonawane 2023). This issue is unacceptable in such a high-risk domain like healthcare, where understanding the reasoning behind a diagnosis is as important as the diagnosis itself. This highlights the need for explainable AI to ensure transparency and increase trust in diagnoses made by AI in medical imaging.

Explainable AI (XAI) refers to a set of processes and methods that make the models results understandable to humans (Narayankar and Baligar 2024). Unlike traditional machine learning models such as linear regression which are more interpretable, deep learning models are more complex and their decision-making processes are typically unclear (Esmaeili et al. 2021). XAI aims to resolve this by providing insights into how these complex models arrive at their predictions. Convolutional Neural Networks (CNNs) are a type of deep learning model that has shown great promise in medical imaging, particularly in detecting tumours from MRI scans (Hosny et al. 2018). While CNNs excel in accuracy, their decision-making process is often opaque. This lack of transparency can be problematic, especially in critical areas like tumour detection, where understanding the model's reasoning is critical (Shin et al. 2023). XAI aims to provide transparency and insight into how these models arrive at their predictions. This is particularly important in the context of healthcare, where the stakes are high, and decisions directly impact patient lives. The integration of XAI techniques with CNNs for brain tumour detection has the potential to significantly improve diagnostic processes and patient care (Esmaeili et al. 2021). By making AI models more interpretable, XAI can enhance the trust and reliability of AI driven decisions in clinical settings.

Despite the advancements in AI and XAI, there remains a gap regarding the development and application of CNN's tailored for explainability in the context of tumour detection from MRI imaging. Many studies focus on the performance and metrics such as accuracy without addressing the crucial aspect of how these models arrive at their decisions (Narayankar and Baligar 2024; Esmaeili et al. 2021). This research aims to fill the gap by developing a CNN designed for brain tumour classification and applying a broad range of XAI techniques to enhance the model's transparency and reliability. The primary research questions that this paper attempts to answer is:

1. How can the integration of multiple XAI techniques enhance the explainability of deep learning models for brain tumour detection compared to single-technique approaches?

2. How does the multiple XAI technique approach perform in explaining model predictions for borderline cases, where 2D MRI slices capture only partial views of the tumour?

This research makes several key contributions to the field of explainable AI in brain tumour detection. Firstly, it provides a comprehensive examination of the current state of XAI techniques in brain tumour detection, offering valuable insights into existing approaches and identifying areas for improvement. Secondly, unlike many studies that rely on pre-trained models this work develops a custom CNN specifically tailored for brain tumour detection. This allows for greater control over the model's architecture and can potentially lead to improved performance and interpretability. Lastly, by integrating multiple XAI methods rather than focusing on a single technique, this study offers a more complete approach to model explainability, enabling a comparative analysis of different XAI techniques and their effectiveness in enhancing the transparency of AI medical diagnostics.

## 2. Literature Review

Medical imaging refers to the various techniques that are used to create visual representations or models of the human body for clinical analysis and medical treatment (An *et al.* 2021; Kumar *et al.* 2023). These methods are generally non-invasive and enable medical professionals to diagnose, monitor and treat medical conditions without the need for exploratory surgery (Selvaraju *et al.* 2017; Zhao *et al.* 2021; Gaur *et al.* 2022; Shin *et al.* 2023). It aids in early detection of diseases, guides surgical procedures, and monitors the progression of diseases and the effectiveness of treatments. Radiology is the medical specialty that focuses on the use of imaging techniques to diagnose and treat diseases (Hosny *et al.* 2018). Radiologists are medical specialists who are trained to interpret medical images and provide essential information to other healthcare providers. The field of radiology has expanded significantly with advancements in imaging technologies, making it integral to the modern healthcare system. Radiologists use a variety of imaging modalities, including X-rays, CT scans, MRI, ultrasound, and nuclear medicine, to obtain detailed images that aid in accurate diagnosis and treatment planning (Najjar 2023; (Ahmed, Asif, *et al.* 2023; Šefčík and Benesova 2023).

Medical imaging has evolved significantly over the years, with various techniques developed to visualise the body's internal structure. X-ray imaging is the oldest form which uses ionising radiation to create two-dimensional images making it excellent for visualising dense structures like bones and teeth but is limited in its soft tissue contrast (Yu *et al.* 2024). Computed Tomography (CT) is an advancement of this technology, providing three-dimensional imaging by using x-rays and computer processing to create cross sectional images of both bone and soft tissue (Najjar 2023). While both x-ray and CT are valuable tools especially in trauma cases and cancer diagnosis, they involve the use of radiation which can lead to health concerns with repeated exposure.

Magnetic Resonance Imaging (MRI) is a sophisticated medical imaging technique that uses powerful magnetic fields, electric fields and radio waves to produce detailed images of the body's internal structures in a non-invasive manner. It provides exceptional imaging of soft tissues like organs and muscle, with superior contrast resolution compared to other imaging techniques (Hussain *et al.* 2022). This makes MRI useful in diagnosing complex neurological conditions (such as brain tumours), musculoskeletal disorders, and cardiovascular diseases. MRI's main strength is its versatility, which stems from its ability to manipulate different parameters in order to highlight different tissue characteristics that allows for a comprehensive evaluation of a patient's anatomy.

MRI offers several different imaging sequences that highlight various different tissue characteristics. The most common sequences are T1-weighted, T2-weighted, and FLAIR (Fluid-Attenuated Inversion Recovery). T1-weighted images provide excellent anatomical detail, with fat appearing bright and fluid appearing dark. T2-weighted images, conversely, show fluid as bright and fat as dark, making them ideal for identifying edema and inflammation. FLAIR sequences suppress the signal from cerebrospinal fluid, enhancing the visibility of lesions and tumours by providing better contrast against surrounding tissues. This makes FLAIR useful for diagnosing lesions and brain tumours (Işın *et al.* 2016; Al-Fakih *et al.* 2024). Despite its many advantages, MRI is characterised by a few substantial limitations. Primarily, its relatively long scan time compared to other imaging modalities can lead to increased costs and patient discomfort. This extended duration also makes MRI more susceptible to motion artifacts which can degrade image quality (Johnson and Drangova 2019). Additionally, the strong magnetic fields used in MRI prevent its use for patients with metallic implants or devices.

To address these challenges ongoing research focusing on developing faster image acquisition techniques and improved image quality. Functional MRI (fMRI) has revolutionised our understanding of brain function by measuring blood oxygenation level-dependent signals. This allows researchers to map brain activity and function which has been useful in studying cognitive processes, neurological conditions and even predicting an individual's behaviour patterns (Kotoula *et al.* 2023). Diffusion Tensor Imaging (DTI) enables the enhanced visualisation of white matter in the brain by measuring the diffusion of water molecules along neural pathways. This has proven useful in researching and providing insights into neurodegenerative diseases and other neurological diseases such as schizophrenia (Podwalski *et al.* 2021). Simultaneous multi-slice imaging or multiband imaging is an advanced imaging technique that excites and records signals from multiple slices simultaneously which reduces scan times significantly. It has also been transformative for DTI and fMRI enabling higher spatial and temporal resolution while maintaining or improving image quality (Barth *et al.* 2016).

AI's integration into medical imaging is gaining momentum, offering both current enhancements in diagnostic capabilities and significant potential for future development in the field of neuroradiology. AI particularly deep learning has been increasingly adopted in medical imaging to assist with tasks such as image segmentation, disease detection, and classification (Najjar 2023). Among the various AI techniques, Convolutional Neural Networks (CNNs) are especially prominent due to their ability to automatically learn features from images. Many studies have shown that AI has the potential to improve diagnostic accuracy, reduce interpretation time, and enhance workflow efficiency in brain tumour imaging. For example, AI algorithms have demonstrated impressive performance in detecting brain tumours in MRI scans (Choudhury *et al.* 2020; Mahmud *et al.* 2023), differentiating between various types of brain tumours (Mzoughi *et al.* 2020; Senan *et al.* 2022; Khaliki and Başarslan 2024), and accurately segmenting tumour regions to aid in detection and classification (Bangalore Yogananda *et al.* 2020). However, despite its promising advancements AI in medical imaging still faces some notable limitations. These include challenges related to data size and quality and lack of explainability with these models (M *et al.* 2024).

Disease detection and classification in medical imaging refers to the process of identifying the presence of abnormalities or pathologies in medical imaging and categorising them into specific disease types or stages (Hosny *et al.* 2018). This enables the early detection, accurate diagnosis and appropriate treatment plan for the patient. AI especially deep learning models have emerged as a powerful tool in this domain, offering the potential to enhance accuracy, speed and consistency in image interpretation (Arabahmadi *et al.* 2022). Recent studies have shown the effectiveness of deep learning models in brain tumour detection. In one study (Mahmud *et al.* 2023) explored the use of a CNN for detecting brain tumours from MRI images. The proposed model achieved a high accuracy showing its promise in early detection of brain tumours. In a similar study by (Choudhury *et al.* 2020) a CNN models was used to detect tumours by classifying them as tumour or non-tumour. The model achieved an accuracy of 96% showing the models potential in identifying tumours at an early stage.

Artificial Intelligence can also be used to classify the type of tumour from MRI images. In a study by (Mzoughi *et al.* 2020) a CNN was designed that classifies MRI images as either low-grade gliomas or high-grade gliomas. This study achieved a high accuracy of 96% benchmarked using the Brats-2018 dataset. In another study (Khaliki and Başarslan 2024) used transfer learning to classify brain tumours in MRI images. The study used transfer learning with a variety of pre-trained models such as VGG19 and InceptionV3 in order to overcome the limited training data and achieved high accuracy in detecting and classifying tumours of different kinds. (Senan *et al.* 2022) performed a similar experiment using a range of pretrained models on the same dataset. They achieved a maximum accuracy of 95% with their AlexNet+SVM model which used a hybrid approach of a pre-trained model and a support vector machine (SVM). This approach showed the strengths of combining a pretrained model with a SVM in order to improve the accuracy of the models' predictions. Segmentation can also aid in the detection and analysis of brain tumours by showing the tumours boundaries and internal structure, helping classify its type. In a study (Bangalore Yogananda *et al.* 2020) developed a fully automated deep learning network for segmenting brain tumours into tumour, tumour core, edema and necrotic tissue. This detailed segmentation enables a more precise understanding of the tumour's characteristics and aids in differentiating between various types of brain tumours, which is critical for planning appropriate treatment strategies and improving patient outcomes.

Despite the promising advancements in AI for brain tumour detection, some notable challenges remain. One significant limitation is the issue of the dataset size and quality. Many studies face challenges related to the limited dataset size, which can lead to overfitting and poor generalisation. For example the studies mentioned above by (Mahmud *et al.* 2023), (Senan *et al.* 2022) and (Khaliki and Başarslan 2024) all included a limited dataset, a dataset that only has 3264 training images. This amount of data is not enough to build a robust model that is representative of the variation in imaging in clinical practice and may lead to model bias. In order to mitigate this the researchers used a variety of pretrained models in order to overcome the limitations of a small dataset in order to achieve high accuracy in their results. A critical challenge in this field is the black box nature of many advanced AI models, especially with Deep Learning (DL). This lack of explainability creates some major barriers to clinical implementation and acceptance (M *et al.* 2024). Deep learning models are often highly accurate and operate in ways that are not easily understandable for humans. They make decisions based on complex patterns in the data that are often at a higher dimension that humans can interpret. This trade-off between interpretability and accuracy are what makes the implementation difficult and can lead to ethical issues (Amann *et al.* 2020). This problem will be discussed in more detail in the next section.

## 2.1. Explainable AI in Medical Imaging

AI particularly deep learning models have shown great promise in medical imaging tasks such as segmentation and disease detection. However, the increasing complexity of these models, especially deep neural networks have led to a significant challenge in terms of interpretability and explainability. This black box problem requires the application of Explainable AI (XAI) techniques in medical imaging. The integration of AI into medical imaging has shown incredible potential in enhancing diagnostic accuracy and efficiency. However, the increasing complexity of AI models, especially Deep Learning (DL) models have given rise to this "black box" problem. This term refers to the lack of transparency in how deep learning models arrive at their predictions (Shin *et al.* 2023). Unlike simpler algorithms where the decision making is clear, deep learning models often make predictions without providing insights into the reasoning behind them. This opacity poses a significant problem in healthcare, where understanding the basis of the prediction or diagnosis is crucial to patient care and clinical decision making (Hosny *et al.* 2018).

DL models especially CNN's have become the most common implementation in medical imaging due to their ability to automatically extract complex features from images. These models are typically composed of multiple layers, with each layer performing some operation on the input data. A typical CNN architecture used for medical image analysis includes convolutional layers that detect features, pooling layers that reduce spatial dimensions, activation layers that introduce non-linearity, and fully connected layers that combine learned features for final predictions. The depth and complexity of these networks, often containing millions of parameters enables them to achieve high accuracy in tasks such as tumour detection or disease classification. However, this same complexity makes it challenging to understand how these models arrive at their decisions (Hosny *et al.* 2018). This challenge highlights a key trade-off in AI development with the balance between model interpretability and accuracy. While simpler models are often more explainable, they may not achieve the high levels of accuracy required for critical medical decisions. On the contrary, complex models like deep neural networks can achieve impressive accuracy but at the cost of reduced interpretability. This is especially troublesome in a medical setting where both accuracy and explainability are crucial for proper patient care.

For brain tumour detection and classification XAI is particularly crucial. Given the significance of a cancer diagnosis, medical professionals need to understand how AI models arrive at their conclusions. XAI can highlight the features of an MRI scan that are the most influential in detecting or classifying a tumour revealing insights that a radiologist.

### 2.1.1. Types of XAI for Brain Tumour Detection

There are several XAI techniques that are useful in providing insights into the decision-making processes of AI models. These techniques aim to make the complex operations of deep learning models more interpretable and understandable to medical professionals. While numerous methods exist, we will focus on three commonly used and powerful XAI techniques: GRAD-CAM, LRP, and SHAP. Each of these methods offers a unique advantage in explaining AI model decisions for brain tumour detection.

**GRAD-CAM**
Gradient-weighted Class Activation Mapping (GRAD-CAM) is a widely used XAI technique that provides a visual explanation for decisions made by a CNN. GRAD-CAM generates heatmaps that highlight the regions in an input image which are most important to the model's decision. The process involves calculating the gradients of the target class with respect to the feature maps of the final convolutional layer. These gradients are then used to compute the weights for the feature maps. These are then combined and passed through an activation function to produce a heatmap, which is then overlayed on the original image in order to produce the heatmap. This displays the regions that have contributed most to the model's decision which enhances transparency (See figure 1) (Selvaraju *et al.* 2017).

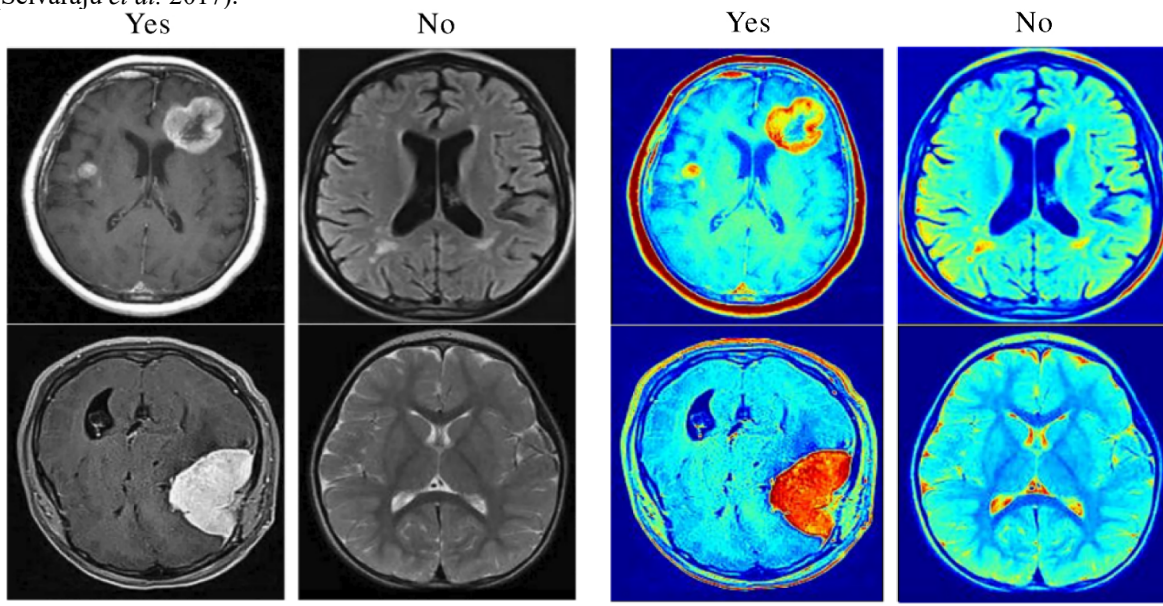

Figure 1: GRAD-CAM showing feature importance (Özbay and Özbay 2023)

Figure 1 shows GRAD-CAM comparing brain MRI scan that contain and do not contain tumours. The left side shows the original MRI scans while the right shows the corresponding GRAD-CAM visualisation. The heatmap highlights regions of interest, where red/orange areas indicate regions that the model considers the most significant for tumour detection, while blue areas are less significant. In the tumour positive cases a distinct red/orange highlights tumour location. In tumour negative cases there is a notable absence of red/orange areas which support a healthy diagnosis. This shows how GRAD-CAM helps to make a model's decision more interpretable.

(Kumar *et al.* 2023) applied GRAD-CAM to a VGG19 model for brain tumour classification, demonstrating its ability to highlight relevant tumour regions in MRI scans. While (M *et al.* 2024) used GRAD-CAM with a ResNet50 model, showing that the generated heatmaps largely highlighted cancerous regions in the MRI images. Similarly (Hussain and Shouno 2023) used GRAD-CAM with a VGG19 and EfficientNet models for multiclass brain tumour classification problem (glioma, meningioma, pituitary, no tumour). (Özbay and Özbay 2023) further extended this approach by combining Grad-CAM with multiple CNN architectures and mRMR feature selection, achieving high accuracy in brain tumour classification while maintaining interpretability. All these studies demonstrated the effectiveness of GRAD-CAM in providing visual explanations for the model's decisions, enhancing the interpretability of AI-driven brain tumour detection. These visualisations could potentially aid radiologists in understanding and validating the AI model's focus areas which would increase trust and transparency of the model.

**SHAP**

SHapely Additive exPlanations (SHAP) is a game-theoretic approach that explains the output of a machine learning model (Zhao *et al.* 2021). SHAP values represent the contribution of each feature to the prediction by comparing what the model predicts with and without the feature. For CNNs applied in medical imaging, SHAP helps in interpreting the contributions of different imaging features, such as texture, intensity, and shape to the AI model's predictions. By breaking down the prediction into multiple individual feature contributions, SHAP allows users to understand how each feature influences the model's confidence in detecting a tumour (Ahmed, Nobel, *et al.* 2023). The main advantage of SHAP in this context is its ability to provide explanations that are both locally accurate (for individual predictions) and globally consistent (across multiple predictions). This means that the sum of the SHAP values for all features equals the difference between the model's prediction and the average prediction, providing a clear and interpretable explanation (Gaur *et al.* 2022).

(Ahmed, Nobel, *et al.* 2023) utilised SHAP in order to increase explainability in their EfficientNetB0 CNN for their multiclass brain tumour problem using MRI images. They presented a unified framework by assigning an explanation of the image for each prediction in order to increase trustworthiness of the prediction. The researchers applied SHAP to provide both local and global explanations of the model's decisions. SHAP values highlighted the most influential regions of the MRI scans for each class prediction, allowing verification of whether the model focused on clinically relevant areas. The study used color-coded pixel attributes to visualise SHAP explanations. Red pixels signified positive SHAP values contributing to a particular class, while blue pixels showed low SHAP values helping to exclude that classification. This approach demonstrated the potential of SHAP to enhance the interpretability in brain tumour detection by AI models (see figure 2).

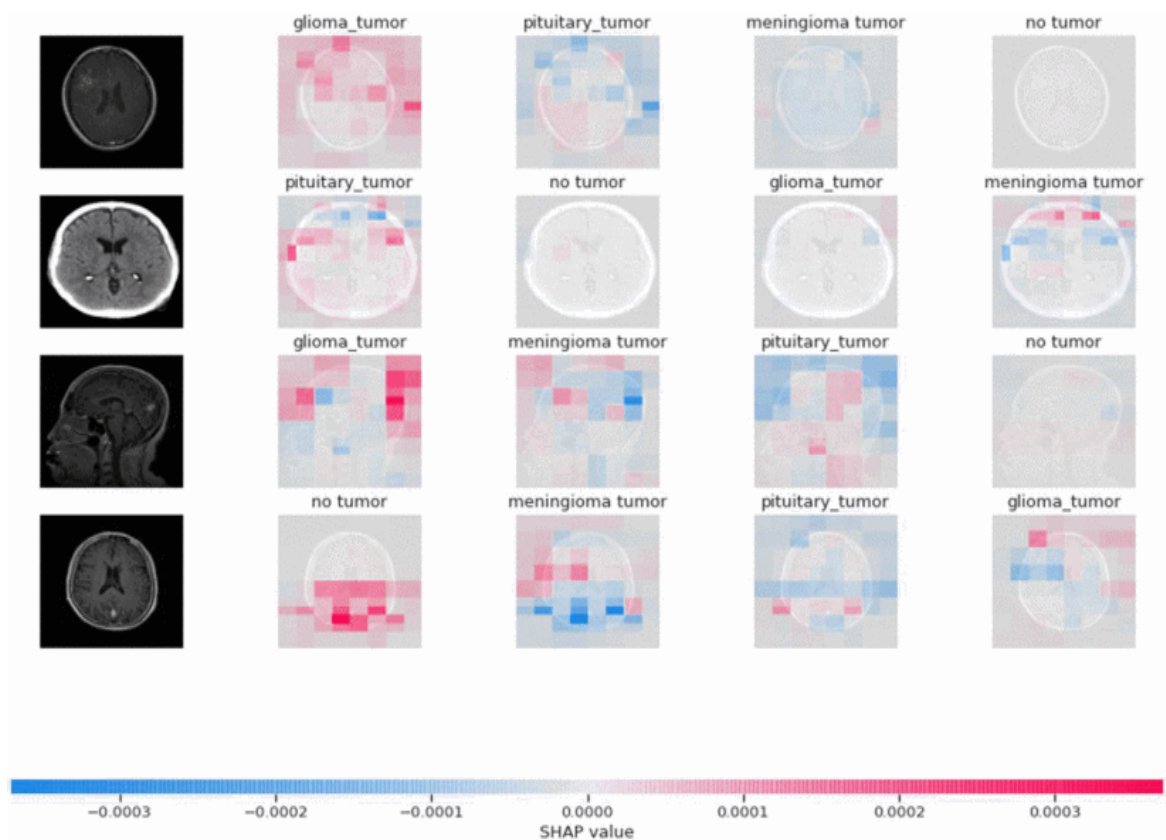

Figure 2: SHAP values (Ahmed, Nobel, et al. 2023)

Figure 2 shows SHAP analysis results for brain tumour MRI classification. Original MRI scans are shown on the left, with their corresponding SHAP value visualisations for different tumour and no tumour cases displayed on the right. The visualisations use a red-blue colour scheme, where red areas indicate regions, the model used to support its classification decision and blue shows regions that influenced against it. This can help provide medical professionals with a means to verify what a model is focusing on in its predictions.

**LRP**

Layer-wise Relevance Propagation is another XAI technique used to explain the decision-making process of AI models. LRP works by propagating the prediction of the model backward through the network layers to the input image, assigning relevance scores to each pixel that contribute to the final prediction. These scores are visualised as heatmaps, similar to GRAD-CAM, which help to highlight the areas of the input that are most critical for the model's decision making (Narayankar and Baligar 2024). (Šefčík and Benesova 2023) applied LRP to explain CNN decisions in their glioma classification model. They trained a CNN model to classify low-grade glioma's (LGG) and high-grade gliomas (HGG) from MRI data. Using LRP they found that their initial model was considering irrelevant features and decided to implement LRP during the training process to force the model to focus on relevant parts of the image by modifying the loss function. This improved the model's accuracy and helped make the model focus on the tumour regions in the MRI images (see figure 3).

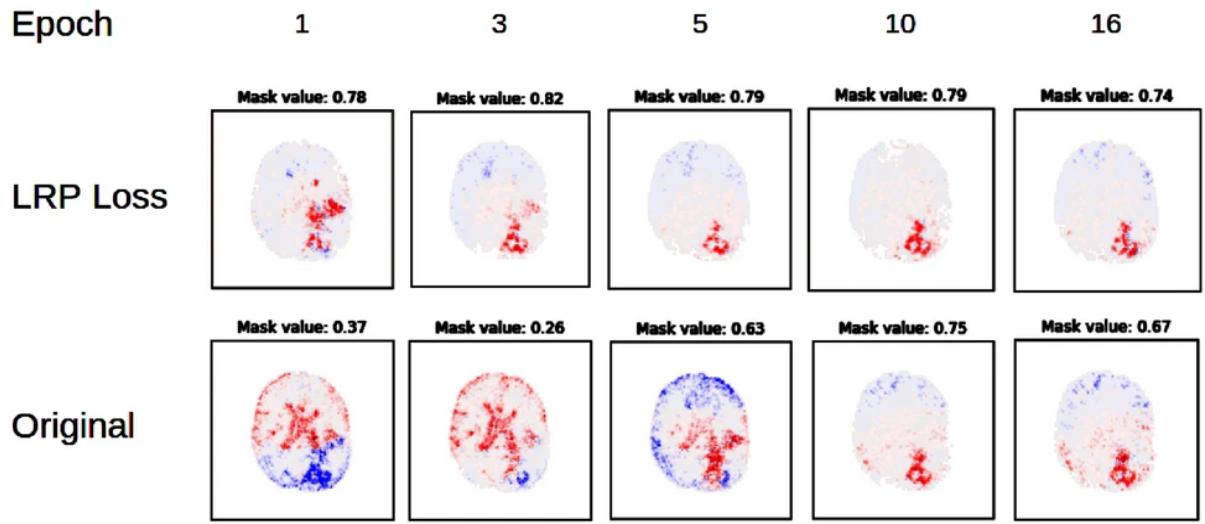

Figure 3: Relevant pixels obtained by LRP during training (Šefčík and Benesova 2023)

In Figure 3, LRP visualisations show how the model learns to focus on tumour regions during training. The top row (LRP Loss) shows the proposed training method achieving focused attention on tumour areas (bright red) early in training, while the bottom row (Original) shows conventional training producing more scattered patterns. The mask values above each area indicate tumour region focus, with higher values showing better focus on clinically relevant areas.

(Ahmed, Asif, *et al.* 2023) used LRP to provide interpretability of a VGG16 model trained to classify brain images as normal/non tumour or tumour. The model achieved high accuracy of 97.33% and the LRP heatmaps revealed which regions of the brain the MRI images were most influential in the model's classification decisions. For normal brain images, the LRP results showed no indications of tumour related features, while for tumour images, the LRP heatmaps highlighted specific regions corresponding to the presence of tumours. Both studies demonstrate how LRP can provide valuable insights into CNN decision-making processes for brain tumour classification, helping to validate the models by showing their focus on clinically relevant brain regions. Additionally, the work by Šefčík and Benesova shows how LRP can be integrated into the training process itself to improve model performance and interpretability. Comparison of XAI Techniques for Brain Tumour Detection

GRAD-CAM, SHAP and LRP have all demonstrated their ability in explaining AI decisions for brain tumour detection each with its own strengths and limitations. These methods differ significantly in their underlying methodologies and the types of explanations they provide as shown in table 1.

| XAI Technique | Strengths | Limitations |
|---|---|---|
| GRAD-CAM | <ul><li>Intuitive visual heatmaps (Selvaraju *et al.* 2017)</li><li>Highlights regions of importance (Kumar *et al.* 2023)</li><li>Easy to understand (Özbay and Özbay 2023)</li></ul> | <ul><li>Limited to the last convolutional layer (Selvaraju *et al.* 2017)</li><li>May miss small details/features (M *et al.* 2024)</li></ul> |
| SHAP | <ul><li>Provides local and global explanations (Ahmed, Nobel, *et al.* 2023)</li><li>Quantifies feature importance (Zhao *et al.* 2021)</li></ul> | <ul><li>Can be computationally expensive (Gaur *et al.* 2022)</li><li>Harder to interpret</li></ul> |
| LRP | <ul><li>Pixel level explanations (Narayankar and Baligar 2024)</li><li>Provides detailed feature</li></ul> | <ul><li>Sensitive to model architecture (Khaliki and Başarslan 2024)</li><li>Computationally expensive for large networks</li></ul> |

Table 1: Strengths and limitations of XAI techniques for brain tumour detection

In terms of interpretability, all three methods produce visual representations that can be understood by medical professionals and patients, but with varying degrees of complexity. GRAD-CAM's heatmaps are often the most intuitive, providing a straightforward visualisation of important regions that is simple for users to understand. LRP's pixel level explanations offer more detailed insights but may require more expertise to interpret. SHAP's feature importance scores can be visualised in various ways, offering flexibility but potentially requiring more work for non-technical users to understand. While most studies in brain tumour detection tend to only apply a single XAI technique, research shows that these methods can be used complimentarily to provide a more comprehensive understanding of a model's decision making. (Narayankar and Baligar 2024) compared multiple XAI techniques for brain tumour analysis using MRI and found that different methods often highlight different aspects of the input image. They suggested that combining multiple XAI techniques could offer a more holistic view of the model's decision-making process.

A key advantage of this multi-method approach is the ability to provide multi-scale interpretations of the model's focus. For instance, using GRAD-CAM to identify broad regions of interest in an MRI scan followed by SHAP or LRP to provide more detailed feature contributions within those regions could offer both a high-level and granular understanding of the model's reasoning. Additionally, the use of multiple XAI techniques can help mitigate the limitations of individual methods. For example, while GRAD-CAM excels at highlighting broad areas of importance, it may miss fine-grained details (M *et al.* 2024). In contrast LRP provides pixel-level explanations but may be more challenging to interpret at a glance (Šefčík and Benesova 2023). By combining these approaches, researchers and clinicians can gain a more balanced and comprehensive view of the AI model's decision-making process.

### 2.3  Current Limitations and Gaps

While XAI techniques have shown promise in enhancing the interpretability of AI models for brain tumour detection, some significant limitations and gaps remain in the current research. One of the most notable gaps in the field is the lack of studies that use multiple XAI techniques to aid explainability in brain tumour detection. While (Narayankar and Baligar 2024) demonstrated the potential benefits of combining XAI methods, most studies such as (Kumar *et al.* 2023), (Šefčík and Benesova 2023) and (Ahmed, Asif, *et al.* 2023) only focus on a single XAI approach. This limits our understanding of how different XAI techniques might complement each other or provide more comprehensive insights into AI model decisions making it easier for people to understand. Current research has not fully explored the potential of XAI in examining false positives (FP) and false negatives (FN) in brain tumour detection models. While many studies report overall accuracy metrics, there is a noticeable lack of in-depth analysis of false positives and false negatives as derived from confusion matrices. This gap represents an opportunity to leverage XAI for understanding, and potentially reducing misclassification rates. XAI techniques can reveal patterns or features that lead to these misclassifications as seen in (Šefčíks and Benesovas 2023) research using LRP. Understanding why an AI model incorrectly classifies certain cases could provide valuable insights for improving model performance and reliability.

# 3. Design

The artifact for this research consisted of six key stages. First the raw three-dimensional (3D) Magnetic Resonance Imaging (MRI) data is converted to two dimensional (2D) slices and pre-processed using Python with NumPy and PIL (Python Imaging Library), applying techniques such as normalisation and augmentation and the data was split using scikit-learn. Next a custom Convolutional Neural Network (CNN) model uses PyTorch for brain tumour detection. The model is trained to classify tumour and non-tumour regions, with the training process optimised using the Adam algorithm. After training, the model evaluation stage assesses the performance of the trained model using various metrics such as accuracy, precision, recall, F1-score, AUC-ROC and confusion matrices. Finally, a range of Explainable AI (XAI) techniques including GRAD-CAM (Gradient-weighted Class Activation Mapping), SHAP (SHapley Additive exPlanations), and LRP (Layer-wise Relevance Propagation) are applied to interpret and visualise the model's decision-making process.

The complete implementation of this project, including all code and documentation, is publicly available on GitHub https://github.com/pmcgon/brain-tumour-xai. This repository contains the preprocessing script and the Jupyter notebook for training the model and performing XAI analysis.

## 3.1. Dataset

We use the Brain Tumour Segmentation (BraTS) 2021 dataset, a famous dataset in the field of medical imaging for brain tumour segmentation (Baid *et al.* 2023). The BraTS 2021 dataset consists of multi-modal MRI images that consist of T1-weighted, T2-weighted, T1-weighted post-contrast (T1ce) and FLAIR sequences. While all these modalities are available in the dataset, this research specifically focused on using FLAIR sequences due to their superior soft tissue contrast for tumour detection. These scans have been manually segmented by experts and categorised into various kinds of brain tumours. While primarily designed for segmentation tasks, this research will adapt the BraTS 2021 dataset for classification, specifically to distinguish between tumour vs non tumour regions in the brain. The BraTS 2021 dataset combines multiple datasets collected from a variety of different institutions (see table 2).

| Collection | Institution/Trial | Cancer Type | No of Subjects |
| --- | --- | --- | --- |
| **ACRIN-FMISO-Brain** | ACRIN 6684 | Glioblastoma multiforme (GBM) | 4 |
| **CPTAC-GBM** | Children's Hospital of Philadelphia | Glioblastoma multiforme (GBM) | 33 |
| **IvyGAP** | Ivy Glioblastoma Atlas Project | Glioblastoma | 30 |
| **TCGA-GBM** | The Cancer Genome Atlas (TCGA) | Glioblastoma multiforme (GBM) | 102 |
| **TCGA-LGG** | The Cancer Genome Atlas (TCGA) | Low-grade glioma (LGG) | 65 |
| **UCSF-PDGBM** | University of California San Francisco | Diffuse Glioma | 263 |
| **UPENN-GBM** | University of Pennsylvania | Glioblastoma | 403 |

Table 2: Collections in BraTS 2021 dataset

The diverse nature of this dataset offers several advantages for building a robust CNN for brain tumour detection. By using data from several different institutions and various kinds of gliomas, the dataset can help to reduce bias and to improve the model's ability to generalise on unseen data. The segmentation was also done manually by experts which provides a high-quality true label for the data which is essential for effective training and validation. The large sample size as a result of combining multiple datasets reduces the risk of overfitting as well as providing diversity among the data. For the artifact associated with this paper, FLAIR (Fluid-Attenuated Inversion Recover) was used as the primary MRI imaging type. FLAIR offers superior soft tissue contrast compared to other MRI sequences which make it effective for highlighting regions where tumours are present (Al-Fakih *et al.* 2024). This characteristic of FLAIR imaging is particularly beneficial for the task of distinguishing between tumorous and non-tumorous regions. Overall, these factors - the diverse dataset, expert segmentation, large sample size, and use of FLAIR imaging - contribute to the development of a CNN that can potentially achieve high performance in distinguishing between tumour and non-tumour regions.

## 3.2. Preprocessing

To prepare the BraTS 2021 dataset for classification, the Nibabel library is used to load and process the raw NIfTI files and their corresponding segmentation masks. 2D slices are extracted from the 3D volumes along the axial (front), axial (top), and sagittal (side) views, with slices selected at fixed intervals (every 5 slices) to reduce redundancy while maintaining representative samples. These extracted slices are then paired with binary labels from the segmentation masks, where slices containing any tumour pixels are labelled as tumorous. This extraction and labelling process is essential for converting the 3D MRI data into a format suitable for 2D classification.

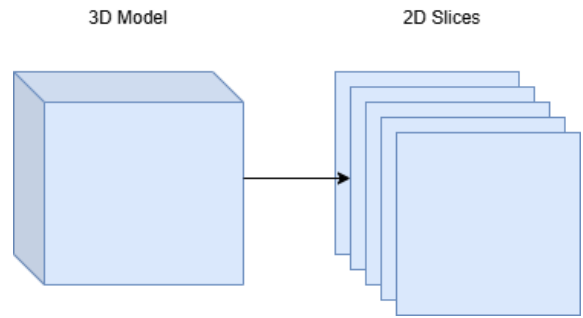

*Figure 4 Converting 3D MRI to 2D Slices*

Additionally, several preprocessing steps are applied to enhance image quality and ensure data consistency. An informativeness assessment function is implemented to retain only slices containing significant brain tissue or tumour regions based on specific thresholds for non-black pixel ratios and edge presence. This step ensures the model is trained only on meaningful and informative data. The selected slices are then normalised to ensure pixel intensities fall between 0 and 1, and each image is resized to 200 x 200 pixels using central cropping to maintain the most relevant features and ensure a consistent input size for the CNN model. For side and front views, images are rotated 90 degrees to ensure natural positioning.

The processed dataset is then split into training, validation and test sets with a ratio of 70:15:15, and split at the subject-level. This means all slices from a single subject are allocated to only one set, preventing any cross-contamination between sets. Initial analysis revealed a class imbalance of ~66% tumour samples across all sets, which was addressed through random under-sampling of the majority class. This reduced the training set from 36,193 to 24,626 samples with 12,313 samples per class. While this reduced the overall dataset size, it ensured a more balanced dataset for training the model. This pre-processing provides the CNN with high-quality, balanced training data while maintaining data independence between sets.

## 3.3. CNN Architecture

Convolutional Neural Networks (CNNs) have demonstrated significant potential in medical imaging analysis particularly for brain tumour detection (Hosny *et al.* 2018). Their ability to automatically extract features from images makes them well-suited for identifying cancerous regions in brain MRI scans. While many studies from the literature review utilised transfer learning approaches with pre-trained models, this project aims to build a custom CNN architecture from scratch. This decision is driven by several factors. Firstly, the BraTS 2021 dataset differs significantly from the natural MRI images used to train most pre-trained models. This is because the images in the BraTS dataset focus solely on brain tissue and do not contain the skull and other features present in a full head MRI. By using a custom CNN, the model can be optimised for this specific data type. Secondly, building a custom model allows for greater control over the architecture, potentially leading to improved interpretability of the model's decision-making process.

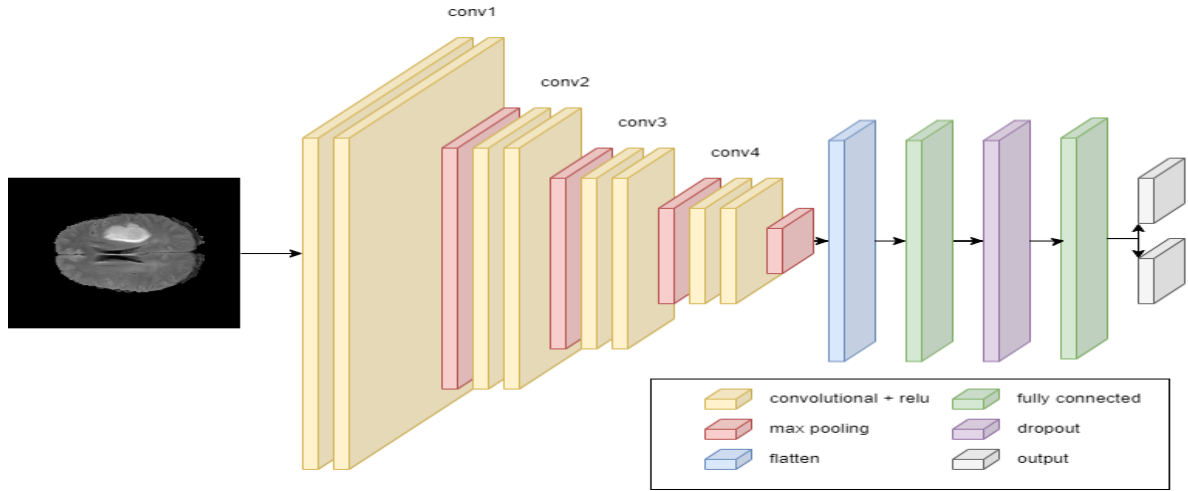

Figure 5: Proposed CNN Architecture

The custom CNN was inspired by (Hafeez *et al.* 2023) who proposed a lightweight CNN model for grading low-grade gliomas and high-grade gliomas using the BraTS 2017 dataset (see figure 4). While their model was developed for a different classification task, its structure provides a strong starting point for the binary classification of tumour vs non-tumour regions especially since it is using a similar dataset. The architecture consists of four convolutional layers paired with max pooling operations, followed by fully connected layers for classification. Each convolutional layer uses increasingly larger kernel sizes (3x3, 5x5, 7x7, and 9x9) and varying numbers of filters (16, 32, 64, and 32) to capture features at different scales. The network concludes with a flatten operation, dense layers with ReLU activation, dropout regularisation (p = 0.5), and a final sigmoid activation for binary classification (see table 3).

| Layer Type | Configuration | Activation | Output Shape | Parameters |
|---|---|---|---|---|
| **Input** | | | (1, 240, 240) | |
| **Conv2D** | 16 filters, 3x3 kernel | ReLU | (16, 238, 238) | 160 |
| **MaxPooling2D** | 2x2 pool size | | (16, 119, 119) | 0 |
| **Conv2D** | 32 filters, 5x5 kernel | ReLU | (32, 115, 115) | 12,832 |
| **MaxPooling2D** | 2x2 pool size | | (32, 57, 57) | 0 |
| **Conv2D** | 64 filters, 7x7 kernel | ReLU | (64, 51, 51) | 100,416 |
| **MaxPooling2D** | 2x2 pool size | | (64, 25, 25) | 0 |
| **Conv2D** | 32 filters, 9x9 kernel | ReLU | (32, 17, 17) | 165,920 |
| **MaxPooling2D** | 2x2 pool size | | (32, 8, 8) | 0 |
| **Flatten Layer** | | | 2048 | 0 |
| **Dense** | | ReLU | 8192 | 16,785,408 |
| **Dropout** | p = 0.5 | | 8192 | 0 |
| **Dense** | | Sigmoid | 1 | 8,193 |

Table 3: Model design

### 3.4. Model Development

The model development is based on the architecture by (Hafeez *et al.* 2023) and was developed using PyTorch. Firstly, a data pipeline was setup. The processed data is loaded using PyTorch's DataLoader class. Data augmentation techniques are applied to the training set using PyTorch's transforms module. These transformations include converting to grayscale, random horizontal flips and random rotations in order to increase the diversity of the training data and potentially improve the model's ability to generalise on unseen data. However, while these augmentations may reduce some biases, they could also introduce new ones if not applied carefully. To mitigate this risk the augmentation techniques were applied conservatively. Extreme transformations such as vertical flips or extreme rotations were avoided in order to maintain the realism of the medical images as seen in previous studies (Choudhury *et al.* 2020; Mahmud *et al.* 2023). This approach ensures that the augmentations did become a limitation; allowing the model to perform more reliably on the data and avoid potential misclassifications from this.

Next the model was created using a custom CNN class from PyTorch's nn.Module. This custom CNN consists of four convolutional layers with ReLU activation, each followed by max pooling and two fully connected layers with a final sigmoid activation as detailed in Table 2. The architecture follows the structure outlined in Table 2. A training function is implemented to handle both the training and validation phases. Binary Cross Entropy Loss will be used as the loss function, and Adam will be used as the optimiser. The Adam optimiser was selected for its adaptive learning rate capabilities, which can lead to faster convergence compared to traditional stochastic gradient descent. The Binary Cross-Entropy Loss function was used due to its suitability for binary classification tasks, effectively measuring the discrepancy between predicted probabilities and actual labels. Following the research by (Hafeez *et al.* 2023), an initial learning rate of 0.0001 with a decay of 0.00002 was used. While these values seem low, were implemented initially to maintain consistency with the reference study. A batch size of 32 was used. This batch size is commonly used in deep learning applications and has been found effective in similar medical imaging tasks (Choudhury et al., 2020; Mahmud et al., 2023). A dropout rate of 0.5 was applied after the first fully connected layer. This means that during each training iteration, 50% of the neurons in this layer were randomly deactivated. This technique helps prevent overfitting by ensuring that the model does not become too reliant on any single neuron. However, the model's performance was closely monitored and these hyperparameters adjusted where necessary to optimise learning.

The training loop initially iterates for 30 epochs. Studies such as (Choudhury *et al.* 2020) used 35 epochs for their custom CNN models. However, the number of epochs can be adjusted based on the model's performance and it could be extended or reduced in order to prevent overfitting. Throughout this process, regular backups of the model are saved to avoid having to retrain the model each time. During each epoch, the training loss and validation loss was calculated using the Binary Cross Entropy loss function which measures the difference between the predicted probabilities and true labels. Training accuracy and validation accuracy are calculated by the ratio of correctly classified samples to the total number of samples. These metrics was recorded and printed throughout the training loop, allowing for real-time monitoring of the model's performance and training progress. After training, the model's performance was evaluated on the test set using various metrics such as accuracy, precision, recall, F1-score, AUC-ROC, and confusion matrices. This provided a comprehensive understanding of the model's performance and its ability to generalise on unseen data.

### 3.5. XAI Integration

After the model was trained and evaluated, GRAD-CAM, SHAP and LRP were used to make the models predications more interpretable. For human interpretation, this integration provides a more comprehensive understanding of the CNN's decision-making process in brain tumour detection. The process begins with the application of each XAI technique independently to the model's predictions. GRAD-CAM generates heatmaps highlighting the region's most important in the model's decision (Selvaraju *et al.* 2017).

LRP was used to backpropagate the relevance scores through the network offering insights into the role of each layer and neuron in the final prediction (Šefčík and Benesova 2023).

SHAP values were calculated to quantify the contribution of each feature (pixel) to the model's output providing a detailed breakdown of feature importance (Zhao *et al.* 2021).

To enhance the interpretability of these XAI techniques, a composite visualisation was created. This consisted of the original MRI image, the GRAD-CAM heatmap, LRP relevance map and SHAP values visualisation (see figure 5).

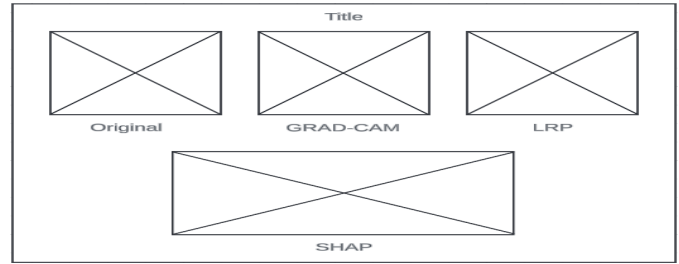

Figure 6: Wireframe for combined XAI image

This combined approach is inspired by (Narayankars and Baligars 2024) work where they compared multiple XAI techniques for classifying different kinds of brain tumours. This method allowed for the quick visual comparison between the different XAI techniques and the original image. This side-by-side presentation enables users to see how each technique highlights different aspects of the image and how they contribute to the model's decision making. This allowed for a better understanding of the model's decision making and help improve trust in the model's capability.

## 4. Results

The original model as proposed by Hafeez et al. (2023) is compared with the improved version of the model, which was fine-tuned during the training process. The improvements to the original model include an extra convolutional layer, a learning rate change from 0.00001 to 0.0001, a learning rate decay from 0.00002 to 0. 0001. Testing showed these changes helped the model learn more effectively and remain stable during training. The batch size was reduced from 128 to 64, resulting in better overall performance. A learning rate scheduler was also added to the training loop to progressively reduce the learning rate over time. This adjustment prevented performance from declining in the later stages of training.

Figure 7 presents the training and validation accuracy for the original model as proposed by Hafeez et al (2023) and the improved version of their model. Figure 8 shows the corresponding loss curves for both models. The accuracy curves demonstrate a clear improvement in performance with the enhanced model. The improved model achieves higher accuracy more quickly and maintains a consistently higher accuracy throughout the training process for both training and validation sets. Starting from around 75% accuracy, the improved model rapidly reaches and keeps its accuracy above 90% for both training and validation data. In contrast the original model shows a more gradual improvement, starting from below 60% accuracy and slowly climbing to around 85% by the end of training.

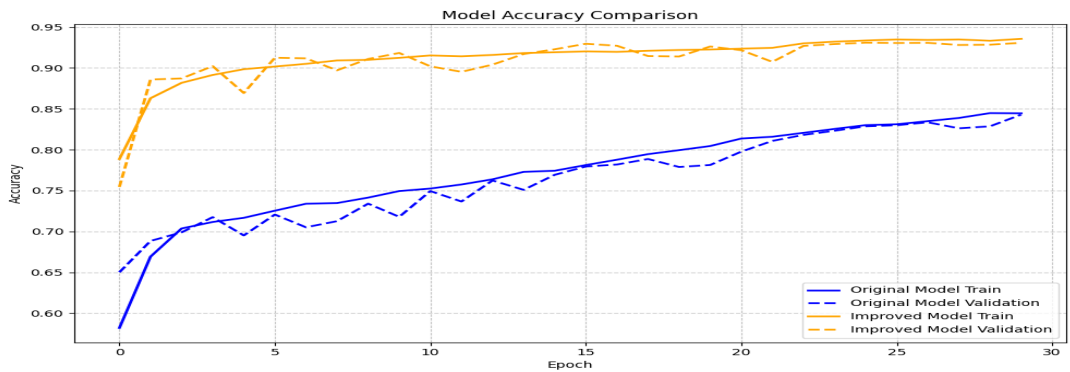

Figure 7: Accuracy learning curves for original and improved model

The loss curves further confirm the better performance of the improved model to the data. It has a much steeper initial decline in loss for both training and validation, and steadily declines and maintains lower loss compared to the original model. By the end of training, the improved model's loss is substantially lower staying at around 0.2, while the original model's loss remains above 0.35.

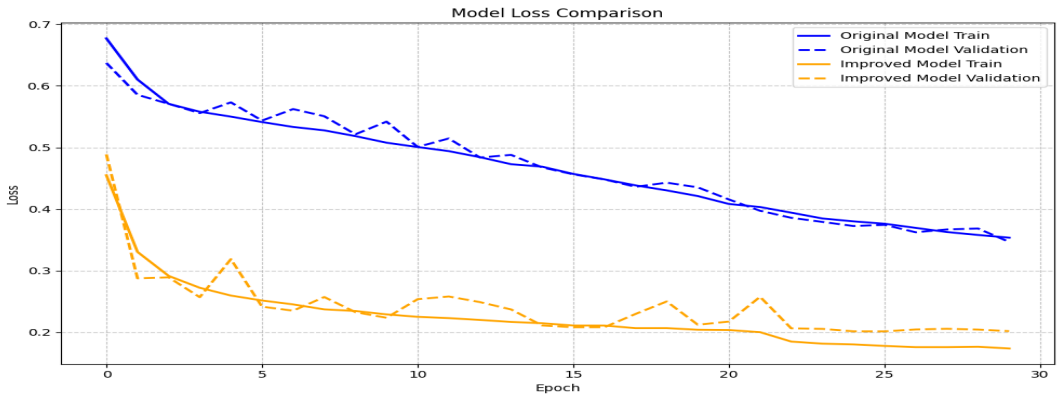

Figure 8: Loss learning curves for original and improved model

Both models show good generalisation, as evidenced by the close tracking of training and validation curves in both accuracy and loss plots. However, it can be seen that the improved model's validation loss shows signs of divergence towards the end of the training, which suggests that the model is potentially beginning to overfit to the training data. The improved model demonstrates better overall performance and faster convergence.

The improved model demonstrates much better performance on the test set when compared to the original model. As shown in Table 4, the test accuracy increased from 0.8476 for the original model to 0.9124 for the improved model. This increase suggests that, within the context of this specific dataset, the enhanced model correctly classifies a higher proportion of brain tumour images. However, it's important to note that individual classifications may still result in false positives or negatives. Similarly, the test loss decreased from 0.3482 to 0.2355. This lower loss suggests that the improved model's predictions are more confident and closer to the true labels. While this reduction indicates improvement, these loss values are still relatively high especially in the context of medical imaging. These metrics show that the architectural changes and hyperparameter adjustments to the improved model have significantly increased its ability to generalise to unseen data. The persisting high loss values despite model improvements, may indicate underlying issues with the dataset itself. The quality of the training and testing data could be limiting factors in the model's performance.

| Model    | Test Accuracy | Test Loss |
|----------|---------------|-----------|
| Original | 0.8476        | 0.3482    |
| Improved | 0.9124        | 0.2355    |

Table 4: Test accuracy and loss for original and improved model

The performance of both the original and improved models can be further analysed through their confusion matrices and other performance metrics as shown in Figure 9. The performance metrics presented in Table 5 demonstrate a clear improvement in the model's capabilities following the improvements.

| Model    | Precision | Recall | F1-Score |
|----------|-----------|--------|----------|
| Original | 0.9191    | 0.7622 | 0.8333   |
| Improved | 0.9608    | 0.8622 | 0.9088   |

Table 5: Precision, recall and f1-scores for the original and improved models

Precision is the ratio of true positives (correctly identified tumours) to the total number of positive predictions (true positives + false positives). It measures the model's ability to avoid labelling non-tumour cases as tumours. The improved model shows an increase in precision from 0.9191 to 0.9608 which is a 4.54% improvement which indicates a reduction in false positive predictions. Recall, also known as sensitivity, is the ratio of true positives to the total number of actual positive cases (true positives + false negatives). It measures the model's ability to find all tumour cases. The recall has improved from 0.7622 to 0.8622, a 13.12% increase, suggesting the enhanced model is significantly better at detecting tumours when they are present which reduces false negatives. The F1-score is the mean of the precision and recall and provides a single score that balances both metrics. The F1-score has increased from 0.8333 to 0.9088, a 9.06% improvement, indicating that the improved model has a better overall balance between minimising false positives and false negatives.

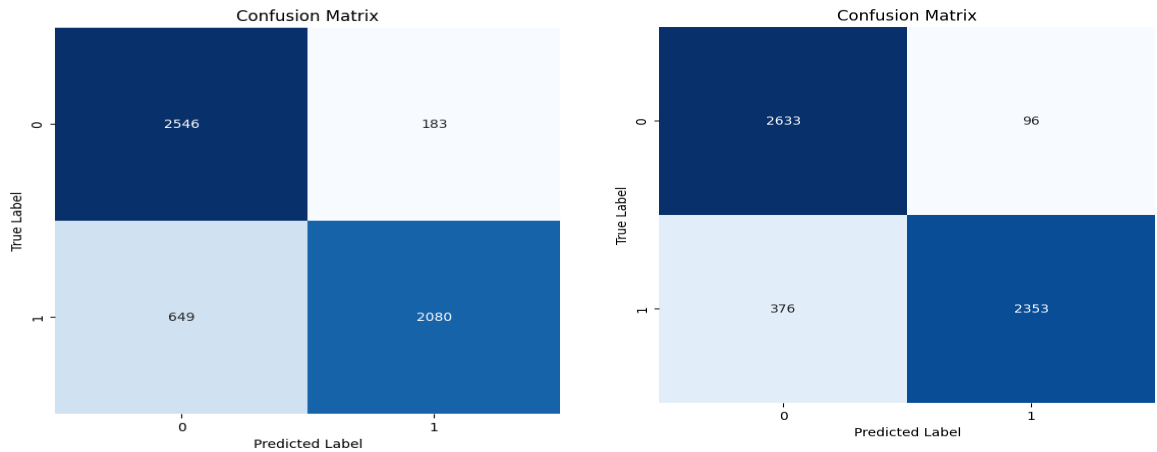

Figure 9: Confusion matrices for the original (left) and improved (right) models

The confusion matrices in Figure 9 provide a more detailed view of the models' performance. The original model (left) shows 2546 true negatives and 2080 true positives, while the improved model (right) demonstrates an increase in both categories with 2633 true negatives and 2353 true positives. This improvement is particularly notable in the correct identification of tumour cases. Furthermore, the improved model shows a reduction in both false positives (96 vs 183) and false negatives (376 vs 649) compared to the original model. The decrease in false negatives is especially significant in a medical context, as it represents fewer missed tumours. These matrices reinforce the performance improvements observed in Table 5 and highlight the improved model's increased accuracy and reliability in tumour detection.

All misclassifications were thoroughly examined to ensure a comprehensive analysis of the model's performance. For the false negatives, 123 out of the 376 false negative images were identified as being of poor quality, with issues such as low contrast, blurriness, and motion artifacts. These factors likely contributed to the model's inability to detect the tumour. For the remaining 253 false negatives which were of appropriate quality, it is assumed that the partial tumours were not prominently visible enough to be accurately detected by the model, leading to these misclassifications.

| Error Type | Count (%) |
| --- | --- |
| Poor Image Quality | 123 (32.7%) |
| Partial Tumours | 253 (67.3%) |

Table 6 False Negative Misclassification Breakdown

These patterns can be seen in Figure 10 where the image on the left suffers from severe blurriness which makes it difficult to detect tumours. Conversely, the image on the right is a much clearer sample however there appears to be no obvious tumour, which suggests that there is very little partial tumour present. The limited visibility of the tumours in these cases may have prevented the model from recognising them as tumour tissue.

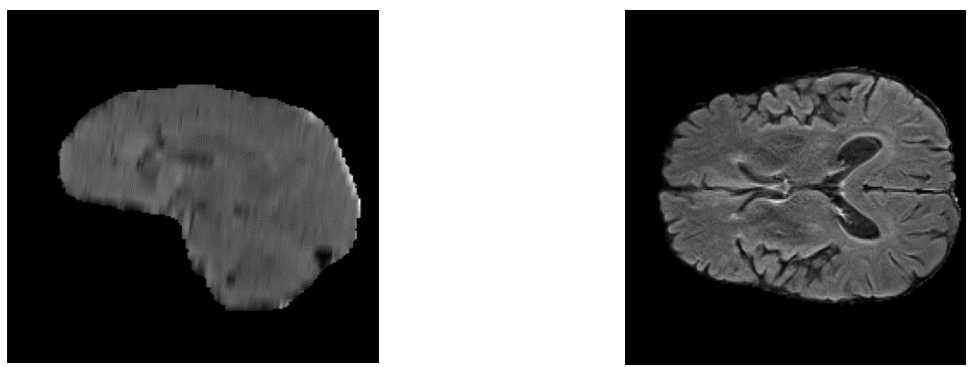

Figure 10: Sample of False Negative Images

Regarding the false positives, 38 out of the total 97 (39.2%) false positive images were found to have poor quality, which may have led the model to incorrectly identify non-tumorous areas as tumours. The majority of false positives (59 cases, 60.8%) were attributed to non-tumorous anomalies that presented imaging characteristics similar to tumorous tissue.

| Error Type | Count (%) |
|---|---|
| Poor Image Quality | 38 (39.2%) |
| Non-tumorous anomalies | 59 (60.8%) |

*Table 7 Falso Positive Misclassification Breakdown*

For example, the image on the left in Figure 11 is severely blurred, which could cause the model to misinterpret noise or unclear regions as tumour tissue. The image on the right, while clearer, shows white spots that could mimic the appearance of a tumour. These spots, which are not part of the segmented tumour (which was used to determine the images class), may be due to other non-tumorous conditions etc. These features likely contributed to the model's incorrect classifications and highlights a significant challenge in distinguishing between actual tumours and other anatomical features that present similar imaging characteristics, suggesting that additional or alternative imaging sequences could potentially reduce the false positive rate.

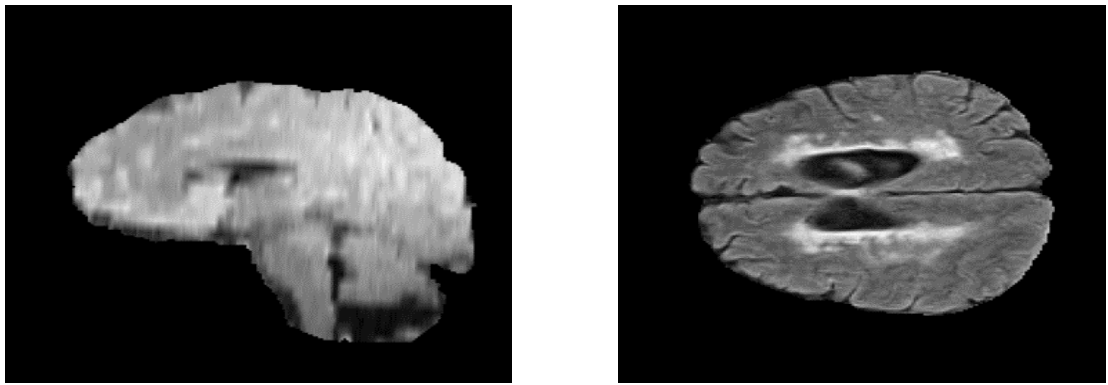

*Figure 11: Sample of False Positive Images*

The ROC curve in Figure 12 provides a visual representation of the performance of the original and improved models ability to distinguish between classes. The improved model which is represented by the orange line achieved an AUC of 0.96 which is much better performance that the original model which has an AUC of 0.92 as shown by the blue line. This indicates that the modifications made to the model have significantly enhanced its accuracy and reliability in detecting tumours, reducing both false positives and false negatives compared to the original model. These improvements were achieved through fine-tuning the models' parameters (as above) and reevaluating the data preprocessing pipeline over several days.

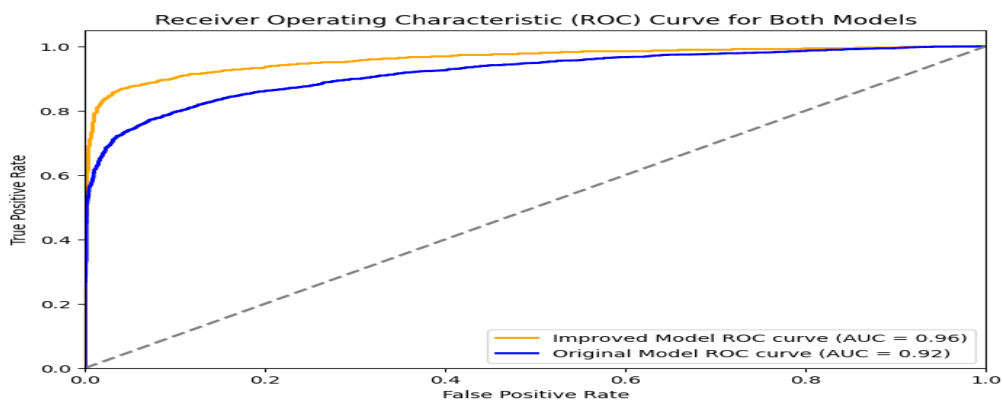

*Figure 12: ROC Curve for the original and improved models*

### 4.1. Explainability Results
GRAD-CAM was applied to visualise the areas in the inputted MRI scans that the model considered most relevant for identifying tumours. The resulting heatmaps highlight the regions in the brain that strongly influence the model's prediction. The layer that was targeted for this was the last convolutional layer as it retains the most spatial information while being the closest convolution layer to the model's final output. Figure 13 shows a

GRAD-CAM visualisation for a positive sample that contains a tumour. The original input image is seen on the left while the GRAD-CAM image is on the right which allows for easy comparison. The GRAD-CAM image shows an intense focus on a specific area within the brain which corresponds to the area where the tumour is on the original image. The heatmap's strong activation around this region, especially the deep blue colours, indicates that the model is highly confident that this particular area is relevant for tumour identification. The precise and concentrated nature of the highlighted area suggests that the model has successfully identified the tumour, focusing its attention where the tumour is most prominent in the scan.

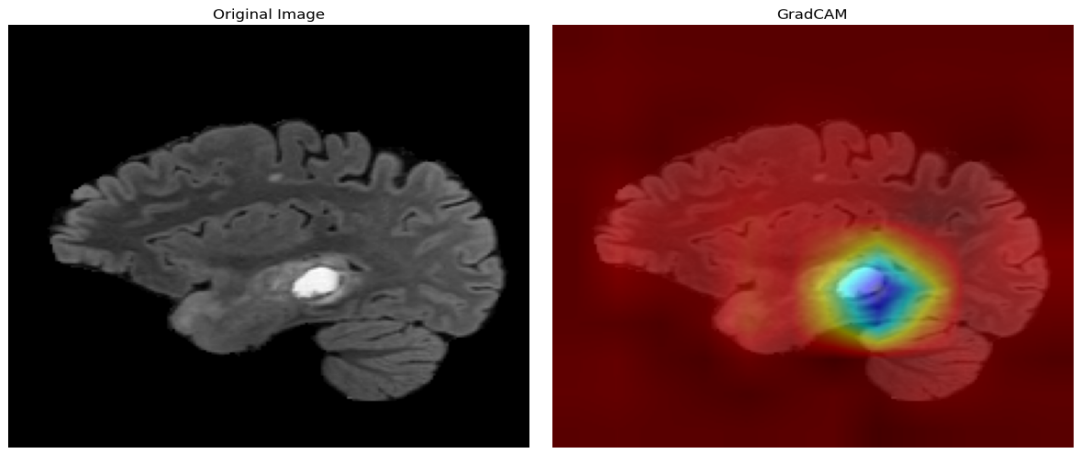

Figure 13: GRAD-CAM visualisation for a positive class (tumour)

Figure 14 presents a GRAD-CAM visualisation for a sample that does not contain a tumour. The original input image is displayed on the left, with the corresponding GRAD-CAM heatmap on the right. The GRAD-CAM image reveals much less concentrated activation pattern across various regions of the brain, with no specific area showing intense focus. This broader and less targeted activation suggests that the model did not identify any region with a strong presence of a tumour. The distribution of attention across the brain indicates that the model is assessing general structural features, but without pinpointing a specific area, which is consistent with a negative classification.

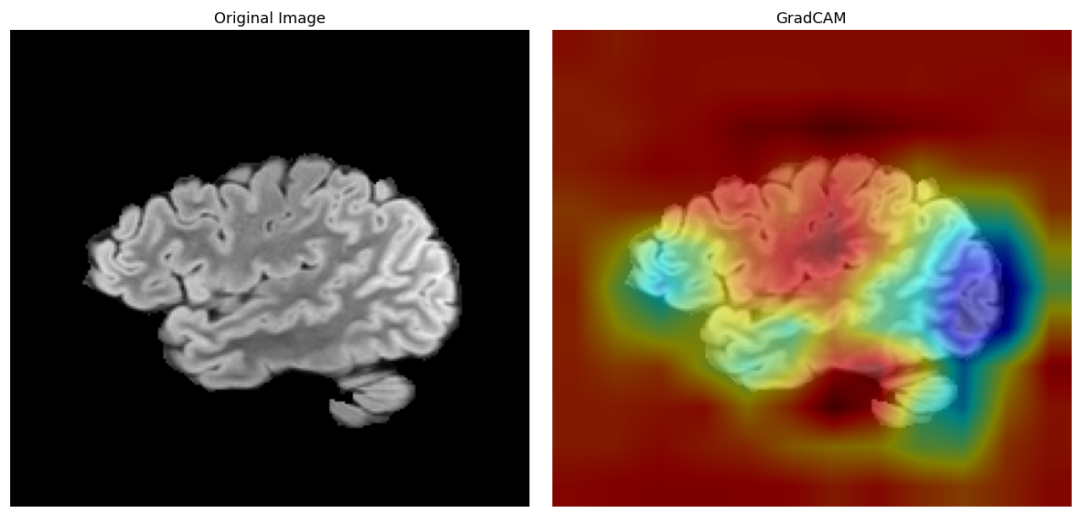

Figure 14: GRAD-CAM visualisation for a negative class (non-tumour)

LRP (Layer-wise Relevance Propagation) was used to visualise the specific regions in the MRI scans that contributed most to the model's tumour predictions. By backpropagating relevance through the network, LRP highlights the pixels that had the greatest impact on the model's decision which provides a detailed view of how the model arrived at its decision.

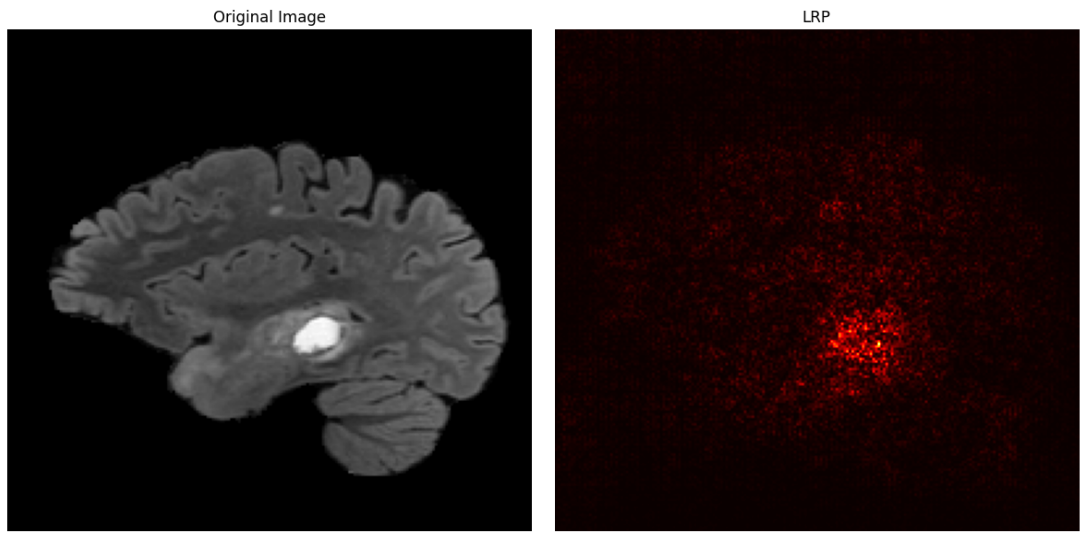

Figure 15: LRP visualisation for a positive sample

Figure 15 shows the LRP (Layer-wise Relevance Propagation) visualisation for a positive sample that contains a tumour. The original MRI image is displayed on the left, while the LRP heatmap is on the right. The LRP visualisation highlights the specific pixels in the image that contributed most significantly to the model's prediction. In this case, there is a concentrated area of high relevance in the centre of the brain which corresponds closely to the location of the tumour seen in the original image. The intense red region in the LRP heatmap indicates that the model relied heavily on this area to make its classification implying that this area's features were key to the model's decision.

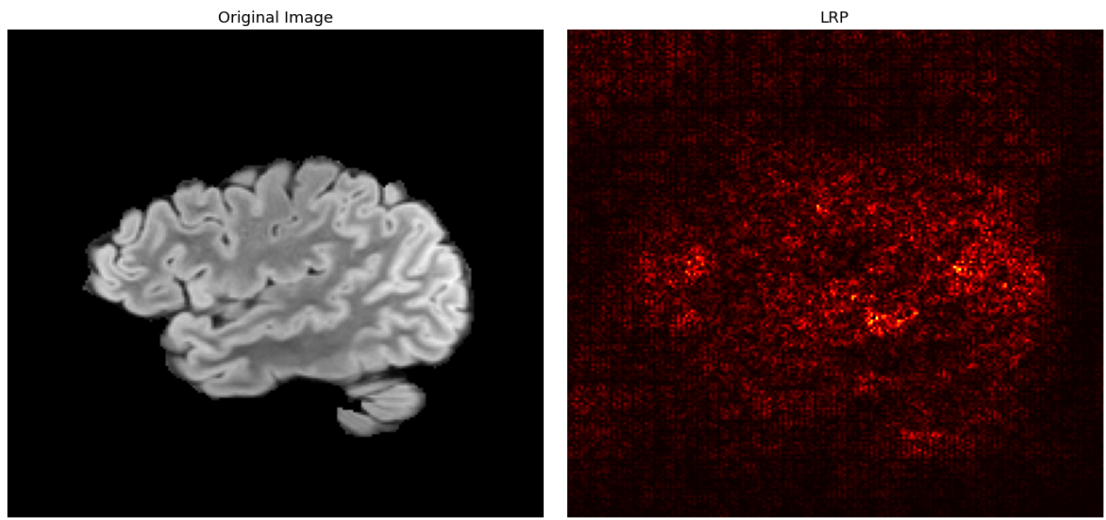

Figure 16: LRP visualisation for a negative sample

Figure 16 contains a LRP visualisation for a sample that does not contain a tumour. Unlike the first image, the LRP visualisation here shows a more distributed pattern of relevance scattered across different areas of the brain. Although there are small clusters of higher relevance, the overall distribution is less concentrated with no single region showing a dominant influence on the model's prediction. This scattered relevance pattern suggests that the model did not find any specific features strongly indicative of a tumour, which aligns with the fact that this sample does not contain a tumour.

SHAP (SHapley Additive exPlanations) was applied to interpret the model's predictions by assigning importance values to different regions of the MRI scans. SHAP values indicate the impact of each pixel on the model's decision, with positive values contributing to the prediction of a tumour and negative values detracting from it.

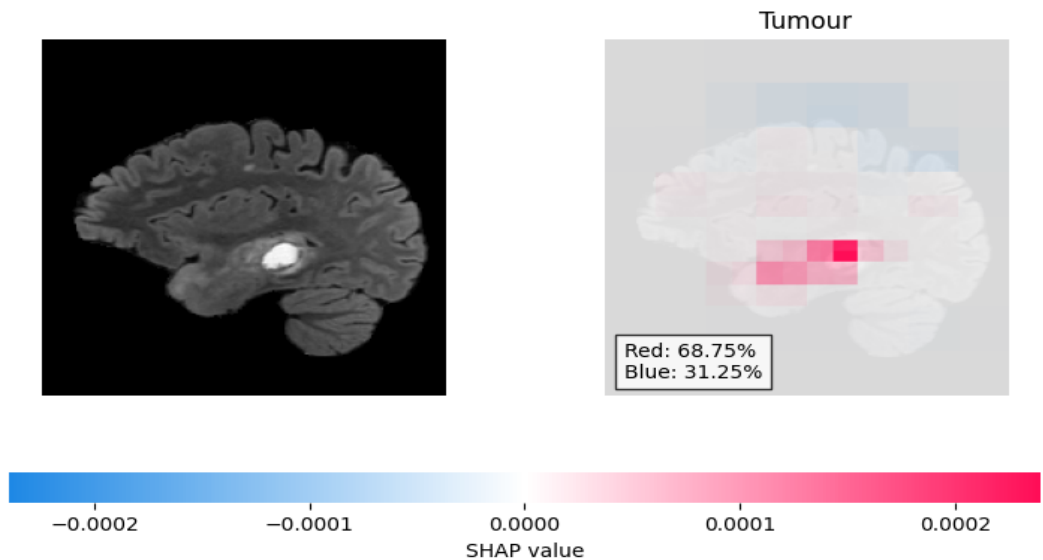

Figure 17: SHAP visualisation for a postive tumour image

Figure 17 shows a SHAP visualisation for a positive sample containing a tumour. The original MRI scan is on the left, and the corresponding SHAP value map is on the right. The SHAP map highlights specific regions of the brain where the model found evidence supporting the presence of a tumour as indicated by the red areas. The concentrated red regions suggest that these areas had a significant positive influence on the model's decision to classify the image as containing a tumour. The blue areas indicate regions that provided evidence against a tumour classification. In this image, 68.75% of the SHAP values contributed positively (red) to the tumour classification while 31.25% contributed negatively (blue).

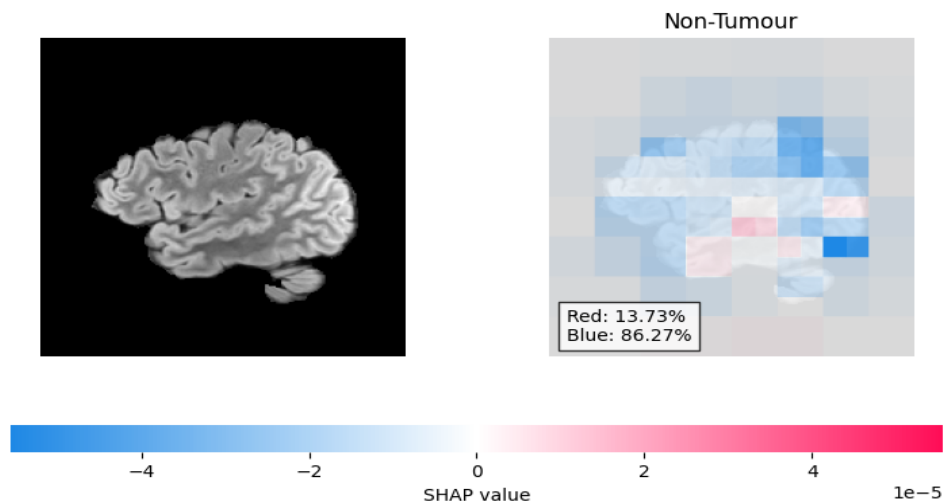

Figure 18: SHAP visualisation for a negative non-tumour sample

Figure 18 shows a SHAP visualisation for a sample that does not contain a tumour. In this SHAP map, blue areas represent regions that contributed to the model's decision to classify the image as non-tumorous, while red areas indicate regions that would have supported a tumour classification. The dominance of blue areas across the brain suggests that the model found more evidence supporting the absence of a tumour, leading to a correct non-tumour classification. The few red areas present had a minimal impact on the overall decision, as the strong negative contributions from the blue regions outweighed them. 13.73% of the SHAP values supported a tumour classification (red), while 86.27% supported a non-tumour classification (blue).

## 4.2 Combined XAI.

After individually exploring GRAD-CAM, LRP, and SHAP, this section focuses on integrating these explainability techniques to achieve a more thorough understanding of the model's predictions. The combination of multiple XAI techniques each providing a distinct yet complementary perspective on the model's decision-making process. GRAD-CAM highlights important spatial regions by highlighting regions of interest at a high level, which makes it useful for the initial localisation of tumour areas LRP offers detailed pixel-level relevance which provides a finer analysis of the pixels within the identified region. SHAP further enhances this by quantifying the contribution of each pixel to the model's decision which provides a numerical basis for understanding the importance of them. When combined, these methods create an interpretation framework where GRAD-CAM identifies the general area of interest, LRP provides detailed feature analysis within that area, and SHAP quantifies the importance of specific features. This layered approach is particularly valuable for detecting partial tumours or analysing borderline cases where single techniques might not be sufficient.

| XAI Technique | Function | Strength | Contribution to Combined Approach |
|---|---|---|---|
| GRAD-CAM | Visualises high level regions | Highlights broad area of interest | Shows high level view of tumour regions |
| LRP | Pixel level analysis | Detailed feature relevance | Fine level analysis of tumour areas |
| SHAP | Quantifies features importance | Quantifies tumour and non-tumour feature contributions | Adds numerical validation to feature significance |

Table 8 Comparison of XAI techniques

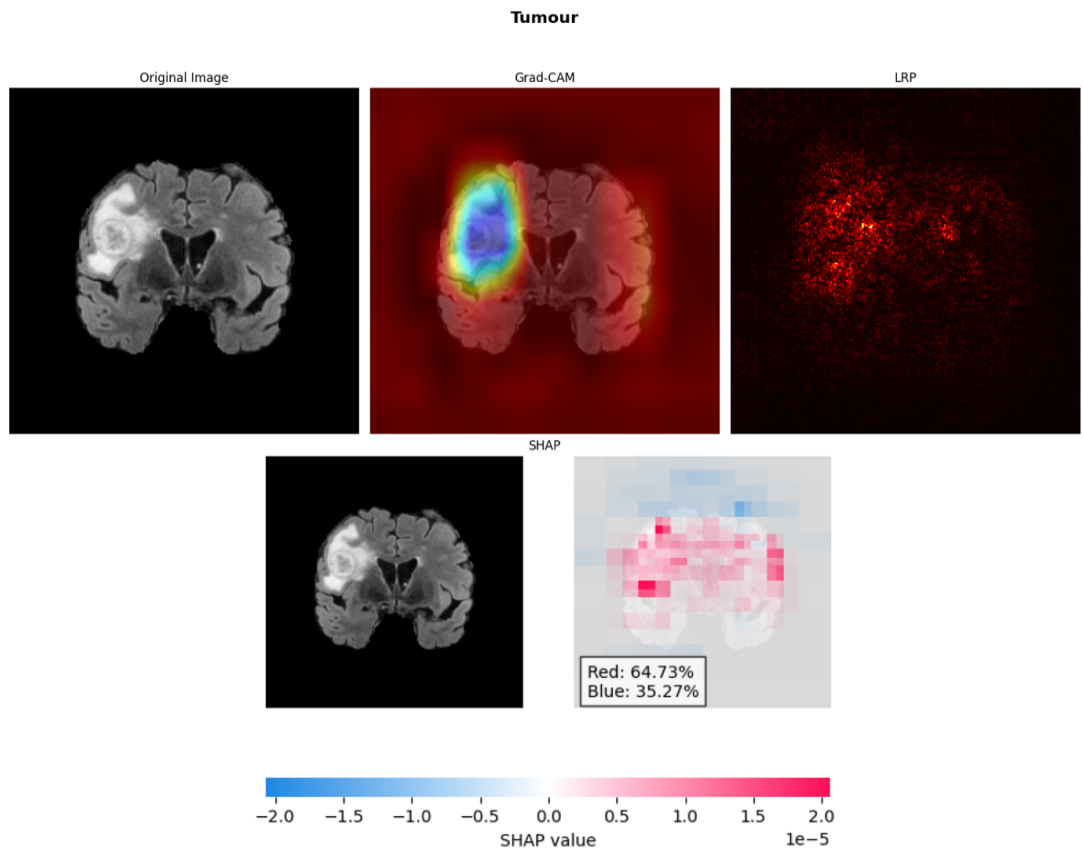

Figure 19: Combined XAI image for a positive sample

In Figure 19, the combined XAI approach provides a detailed view of the model's decision making. The original image on the upper left shows a clear view of the tumour which appears as a bright white area. The GRAD-CAM heatmap highlights a specific area in the brain where the model concentrated its attention, corresponding directly to the tumour visible in the original scan. LRP further supports this by showing a concentrated relevance cluster in the same area, indicating that this region was crucial in the model's classification of the image as tumorous. SHAP values add another layer of understanding, with strong red areas pinpointing the features that contributed most to the tumour classification, while blue areas played a minimal role. In this SHAP image there is a mix of light pink to red areas spread across much of the brain, including but not limited to the tumour region with 64.73% of the SHAP values supporting a tumour classification (red) while 35.27% contributed against it (blue). This suggests that while the tumour area contributes positively to the classification, the model is also considering a broader set of features across the entire image. However, it's crucial to note that even with these advanced XAI techniques, false negatives or false positives are still possible. This highlights the importance of using these XAI methods in conjunction with human expertise. These visualisations should be used to supplement medical diagnoses to ensure an accurate diagnosis. This consistency across GRAD-CAM, LRP and SHAP shows how the model effectively identified the tumour.

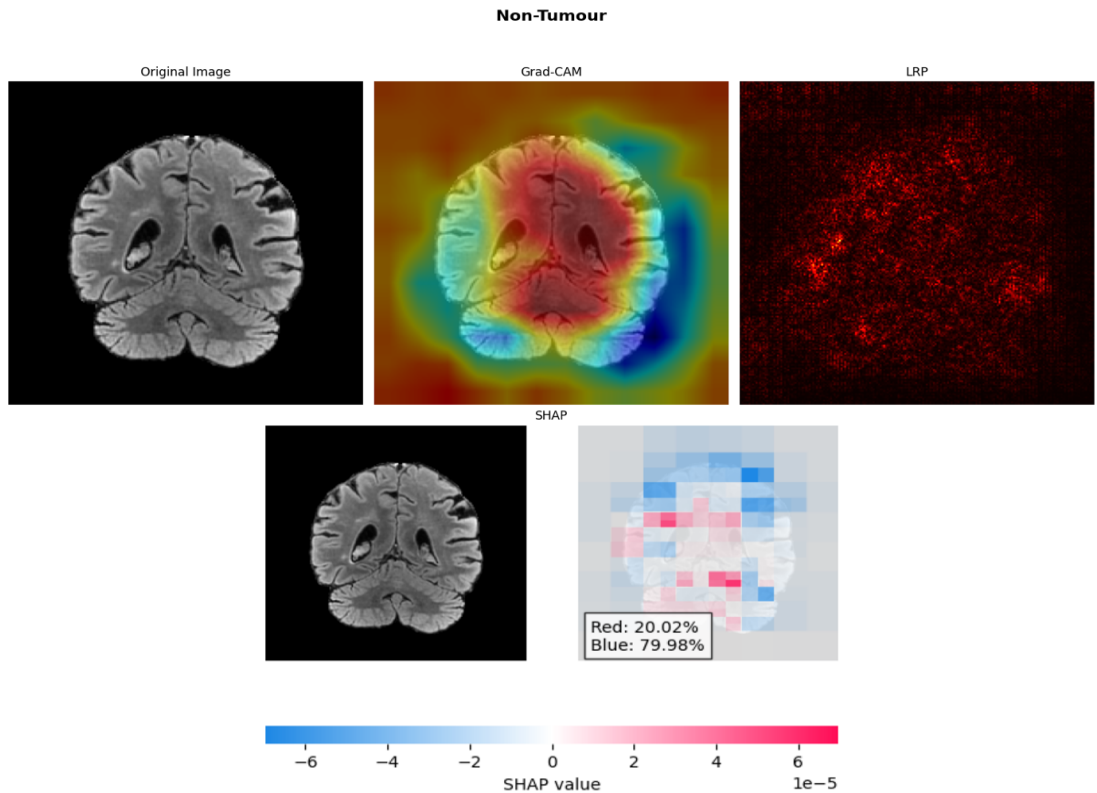

Figure 20: Combined XAI image for a negative class sample

Figure 20 contains the combined XAI approach for a non-tumour sample. The GRAD-CAM heatmap shows the models broad area of attention across the brain which focuses on certain areas without pinpointing any specific area. The LRP visualisation indicates scattered relevance across the image, with no intense focus which indicates that no single region was strongly influential in the model's decision. The SHAP values reinforce this interpretation, with 20.02% of the SHAP values supported a tumour classification (red), while 79.98% contributed against it (blue).

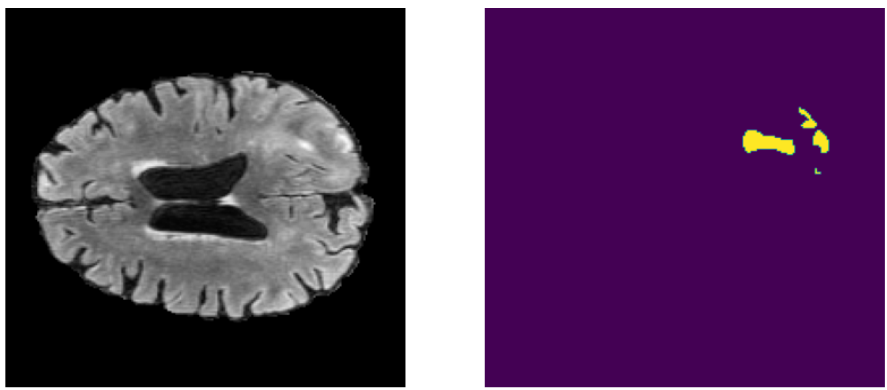

Figure 21: Original image (left) and segmentation mask (right)

The combined XAI approach was tested on a partial tumour to evaluate the model's ability to detect and interpret subtle tumour features. In this case, the original image shows an image that contains a partial tumour which is confirmed by the segmentation mask as seen in Figure 21. The model's predictions were then analysed using GRAD-CAM, LRP, and SHAP to gain a comprehensive understanding of how it identified this partial tumour.

Figure 22 contains the combined approach for the partial tumour. Unlike the previous approaches the original image and the segmentation mask were combined to create a guide for viewers to follow. The GRAD-CAM heatmap reveals that while the model's attention is focused on the region where the partial tumour is located, it also highlights other areas of the brain. This broader focus might indicate some uncertainty in the model's decision-making, suggesting that the model considers multiple regions when classifying the image which could be due to the partially developed nature of the tumour. While GRAD-CAM alone might suggest uncertainty, the LRP visualisation provides a more focused image, showing that the model assigns relevance primarily to the region where the tumour is located. This pixel level detail helps confirm the model's detection of the tumour features despite its partial appearance. SHAP shows both positive (red) and negative (blue) contributions. In this case, 67.48% of the SHAP values supported a tumour classification, while 32.52% contributed against it. While the model shows promise in detecting some partial tumours, its performance can vary depending on the specific characteristics of each image. This combined XAI approach proves particularly valuable in such borderline cases, where individual XAI techniques alone might provide ambiguous results

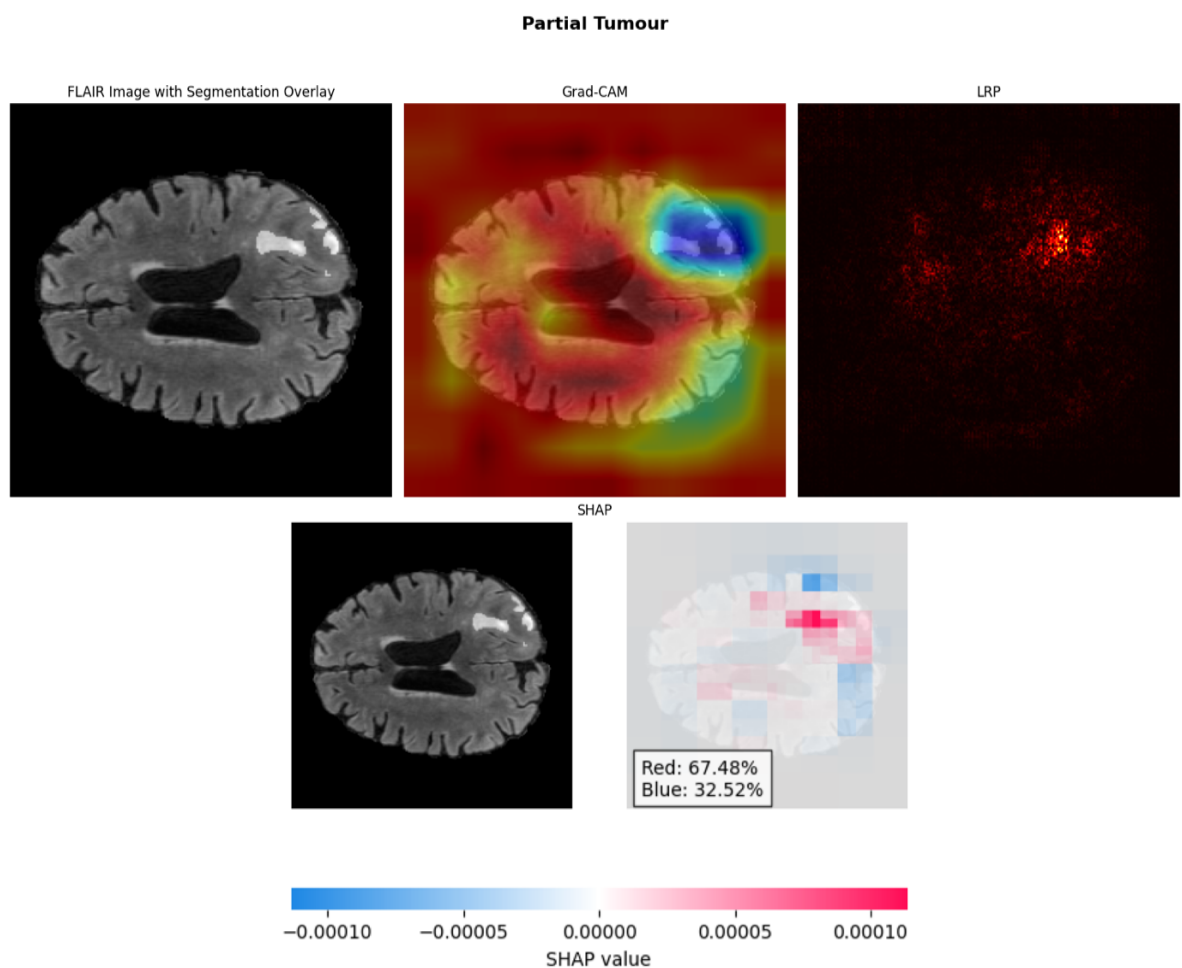

Figure 22: Combined XAI approach on partial tumour

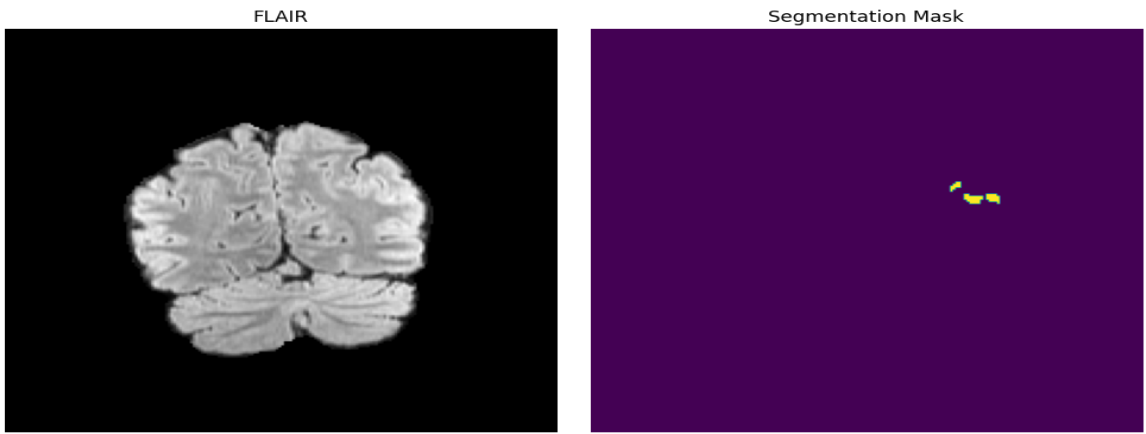

Figure 23 Original image (left) and segmentation mask (right)

A second partial tumour case as seen in Figure 23 helps demonstrate how multiple XAI techniques provide insight into challenging tumour presentations. In this case while the original image shows minimal tumour visibility the segmentation mask confirms the presence of tumorous tissue. In this scenario the model predicted a tumour with a probability of 0.507 indicating significant uncertainty. This is reflected in the combined XAI visualisations seen in Figure 24. The GRAD-CAM heatmap shows multiple areas of activation which reflects the uncertainty in the tumour's localisation. LRP shows scattered relevance rather than a concentrated focus which aligns with a non-tumour diagnosis. SHAP quantifies this uncertainty with 85.97% of values contributing to a non-tumour diagnosis vs 14.03% for a tumour diagnosis. This combination of XAI techniques highlights the model's uncertainty in subtle cases emphasising the need for additional scrutiny by medical professionals. In such cases examining contiguous MRI slices could provide more comprehensive insights and aid in decision-making.

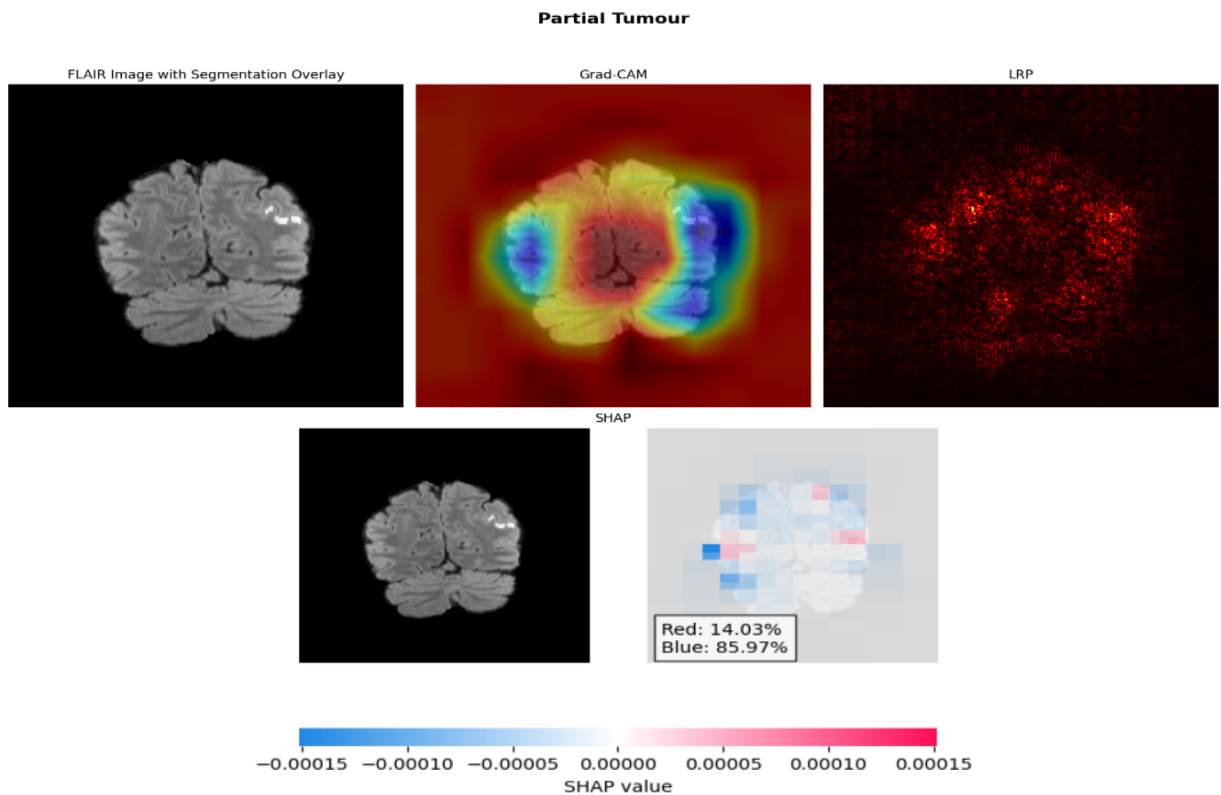

*Figure 24* Combined XAI approach on partial tumour

Each XAI method presents distinct challenges when used separately. GRAD-CAMS's broad heatmap can sometimes highlight areas beyond the actual tumour region as seen in the partial cases above. LRP while providing detailed pixel level analysis can sometimes show scattered patterns that require careful interpretation to distinguish between relevant tumour features and normal tissue. SHAP provides valuable quantitative insights, but it can be difficult to interpret especially if used on its own and without qualitative information from the other XAI methods.

These limitations could possibly hinder the reliability of XAI in clinical applications, where precise and interpretable explanations are needed. The combined approach mitigates these challenges by integrating GRAD-CAM's regional focus with LRP's pixel-level precision and SHAP's quantitative insights, offering a comprehensive explanation that validates or challenges single-method interpretations.

## 4.3 Summary

The improved model demonstrated superior performance over the original model, with test accuracy increasing from 84.76% to 91.24% and test loss decreasing from 0.3482 to 0.2355. Learning curves revealed faster convergence and consistently higher accuracy throughout training for the improved model, indicating better overall performance and generalisation capabilities. Precision, recall, and F1-score metrics all showed substantial improvements. Precision increased by 4.54% (from 0.9191 to 0.9608), recall a 13.12% increase (from 0.7622 to 0.8622), and the F1-score improved by 9.06% (from 0.8333 to 0.9088). Confusion matrix analysis revealed improvements in classification accuracy. These changes demonstrate the improved model's enhanced ability to correctly identify both tumour and non-tumour cases, significantly reducing misclassifications. Misclassification analysis provided insights into the sources of remaining errors. From a sample of 100 false negatives (FN), 44% were caused by poor image quality, while 56% were likely due to partial tumours or limited visibility. Among a sample of 50 false positives (FP), 40% were linked to poor image quality, with others potentially resulting from other medical anomalies. This analysis highlights areas for potential further improvement in data quality and more robust preprocessing steps.

The application of XAI techniques offered insights into the model's decision-making process. GRAD-CAM demonstrated focused attention on tumour regions in positive samples and distributed attention in negative samples. LRP provided precise pixel-level relevance, highlighting tumour areas in positive samples and showing scattered relevance in negative samples. SHAP quantified feature contributions, effectively distinguishing between tumour and non-tumour regions with color-coded visualisations. Each technique contributed unique perspectives, enhancing the overall interpretability of the model. The combined XAI approach of GRAD-CAM, LRP and SHAP provided a comprehensive interpretability framework. This approach successfully identified partial tumours, demonstrating the model's capability to detect subtle features. It offered a deeper understanding of model decisions which increases trust in the model's decision-making. However, cognisance should be given to some key limitation identified from these results. These include poor image quality affecting classification accuracy, challenges in detecting partial or less visible tumours, potential misinterpretations due to non-tumorous anomalies and model uncertainty in borderline cases.

## 4.4 Clinical Implications and Limitations

The combined XAI framework demonstrates significant potential as a Decision Support Tool (DST) in clinical practice, while also presenting important limitations that require consideration. While the underlying model's classification accuracy (91.24%) indicates more development would be needed for clinical deployment, the primary contribution of this work lies in demonstrating how multiple XAI techniques can provide complementary insights into model decision-making.

As a DST, the framework offers valuable support to medical professional's diagnostic processes without attempting to replace clinical expertise. The multi-layered explanations provided by GRAD-CAM, LRP, and SHAP can complement radiologists' expert interpretation by offering different perspectives on regions of interest in the MRI scans. This is particularly valuable in cases of subtle or partial tumours, where the model's attention to specific regions could provide additional insights that align with or enhance the medical professionals' assessment. By presenting these visual explanations in different forms, from broad regional highlighting to detailed feature mapping, the framework supports the diagnostic process of medical professional. Furthermore, the system could serve as a quality assurance mechanism by highlighting areas that might warrant additional attention.

However, several significant limitations impact the framework's current clinical usability. The model's reliance solely on FLAIR imaging sequences, while useful for highlighting certain tumour characteristics does not reflect clinical practice where radiologists typically analyse multiple MRI sequences (T1, T2, T1-contrast) to make comprehensive diagnoses. The framework's current restriction to 2D slice analysis is particularly limiting, as clinicians typically examine sequences of contiguous MRI slices to build a comprehensive 3D understanding of tumour extent and characteristics. Medical professionals may also find the framework too complex especially if multiple XAI techniques are combined. Furthermore, integrating the framework into existing clinical workflows may be challenging due to factors such as workflow disruption and technical resource requirements which could limit the framework.

## 5 Conclusion

This study set out to explore how multiple Explainable AI (XAI) techniques could be integrated to enhance the interpretability of deep learning models for brain tumour detection. Specifically, the research aimed to answer how can the integration of multiple XAI techniques enhance the explainability of deep learning models for brain tumour detection compared to single-technique approaches?

The integration of GRAD-CAM, SHAP, and LRP significantly enhance the explainability of deep learning models used for brain tumour detection, with each technique offering a unique insight into the model's decision-making process. This multi-technique approach supports the findings of (Narayankar and Baligar 2024) who suggested that combining multiple XAI techniques could offer a more holistic view of a model's decision-making process GRAD-CAM highlights the regions of the brain that the model considers most relevant. This aligns with the work of (Kumar *et al.* 2023) who demonstrated GRAD-CAM's effectiveness in highlighting relevant tumour regions in MRI scans. LRP provides a pixel-level analysis of which areas contributed the most to the model's predictions. This supports the findings of (Šefčík and Benesova 2023) who showed how LRP could reveal the model's focus on specific tumour regions. SHAP quantifies the impact of individual features on the model's output, similar to the approach used by (Ahmed, Nobel, *et al.* 2023) in their multiclass brain tumour classification study.

By combining these techniques, a much more comprehensive look into the model's decision-making process is available compared to using any single method. This approach can help increase trust in and help deepen the understanding of the models' classifications. The multiple XAI approach demonstrated good ability in explaining the model predictions for borderline cases where 2D MRI slices capture only partial views of tumours. In cases of partial tumour visibility, GRAD-CAM visualisations often show a broader focus across multiple brain regions, indicating model uncertainty in these challenging scenarios. LRP provides more focused visualisations highlighting the model's ability to assign relevance primarily to regions where partial tumours are located, even when they're not fully visible. SHAP further enhances this interpretation by quantifying the importance of different features, sometimes revealing a significant proportion of values supporting tumour classification despite incomplete visibility. The use of multiple techniques provides a more reliable framework for examining partial tumours.